\documentclass{article}

\pdfoutput=1

\PassOptionsToPackage{numbers, compress}{natbib}

\usepackage[final]{neurips_2021}

\usepackage[utf8]{inputenc} 
\usepackage[T1]{fontenc}    
\usepackage{hyperref}       
\usepackage{url}            
\usepackage{booktabs}       
\usepackage{amsfonts}       
\usepackage{nicefrac}       
\usepackage{microtype}      
\usepackage{amsmath,amsfonts,amssymb, amsthm}
\usepackage{bm}
\usepackage{subcaption}
\usepackage{hyperref}
\usepackage{url}
\usepackage{graphicx}
\usepackage{wrapfig}
\usepackage{physics}
\usepackage{placeins}
\usepackage{xcolor}
\usepackage{float}
\usepackage{flafter}
\usepackage{mathtools}
\usepackage{multirow}
\usepackage{bbold}

\title{Learning Compact Representations of Neural Networks using DiscriminAtive Masking (DAM)}

\author{%
   Jie Bu$^*$\\
   Virginia Tech\\
   \texttt{jayroxis@vt.edu}
   \And
   Arka Daw$^*$\\
   Virginia Tech \\
   \texttt{darka@vt.edu} \\
   \AND
   M. Maruf$^*$\\
   Virginia Tech\\
   \texttt{marufm@vt.edu} \\
   \And
   Anuj Karpatne\\
   Virginia Tech \\
   \texttt{karpatne@vt.edu} \\

}
\makeatletter

\newcommand{\diag}{\textup{diag}}
\newcommand{\thickhline}{%
    \noalign {\ifnum 0=`}\fi \hrule height 0.65pt
    \futurelet \reserved@a \@xhline
}

\newtheorem{theorem}{Theorem}

\DeclarePairedDelimiter{\ceil}{\lceil}{\rceil}
\makeatother

\begin{document}
\maketitle
\def\thefootnote{*}\footnotetext{These authors contributed equally to this work.}

\begin{abstract}


A central goal in deep learning is to learn  compact representations of features at every layer of a neural network, which is useful for both unsupervised representation learning  and structured network pruning. While there is a growing body of work in structured pruning, current state-of-the-art methods  suffer from two key limitations: (i) instability during training, and (ii) need for an additional step of fine-tuning, which is resource-intensive. At the core of these limitations is the lack of a systematic approach that jointly prunes and refines weights during training in a single stage, and does not require any fine-tuning upon convergence to achieve state-of-the-art performance. 
We present a novel single-stage structured pruning method termed DiscriminAtive Masking (DAM). The key intuition behind DAM is to discriminatively prefer some of the neurons to be refined during the training process, while gradually masking out other neurons. 
We show that our proposed DAM approach has remarkably good performance over a diverse range of applications in representation learning and structured pruning, including dimensionality reduction, recommendation system, graph representation learning, and structured pruning for image classification. We also theoretically show that the learning objective of DAM is directly related to minimizing the $L_0$ norm of the masking layer. 
All of our codes and datasets are available \url{https://github.com/jayroxis/dam-pytorch}.


\end{abstract}

\section{Introduction}




A central goal in deep learning is to learn  \textit{compact} (or sparse) representations of features at every layer of a neural network that are useful in a variety of machine learning tasks. For example, in unsupervised \textit{representation learning} problems \cite{bengio2013representation}, there is a long-standing goal to learn low-dimensional embeddings of input features that are capable of reconstructing the original data \cite{autoencoder,ranzato2007sparse,le2011optimization,olshausen1996emergence}. 
Similarly, in supervised learning problems, there is a growing body of work in the area of \textit{network pruning} \cite{blalock2020state}, where the goal is to reduce the size of 
modern-day neural networks (that are known to be heavily over-parameterized \cite{denil2013predicting, zhu2018improving, kornblith2019similarity, wang2020orthogonal}) so that they can be deployed in resource-constrained environments (e.g., over mobile devices) without compromising on their accuracy. 
From this unified view of representation learning and network pruning, the generic problem of ``learning compact representations'' has applications in several machine learning use-cases such as dimensionality reduction, graph representation learning, matrix factorization, and image classification.


A theoretically appealing approach for learning compact representations is to introduce regularization penalties in the learning objective of deep learning that enforce $L_0$ sparsity of the network parameters, $\theta$. However, directly minimizing the $L_0$ norm requires performing a combinatorial search over all possible subsets of weights in $\theta$, which is computationally intractable. In practice, a common approach for enforcing sparsity is to use a continuous approximation of the $L_0$ penalty in the learning objective, e.g., $L_1$-based regularization (or Lasso \cite{tibshirani1996regression}) and its variants \cite{seto2021halo,yun2019trimming}. While such techniques are capable of pruning individual weights and thus reducing storage requirements, they do not offer any direct gains in inference speed since the number of features generated at the hidden layers can still be large even though the network connectivity is sparse \cite{blalock2020state}. Instead, we are interested in the area of \textit{structured network pruning} for learning compact representations, where the sparsity is induced at the level of neurons by pruning features (or channels) instead of individual weights.


\begin{wrapfigure}{t}{0.33\textwidth}

\includegraphics[width=0.33\textwidth]{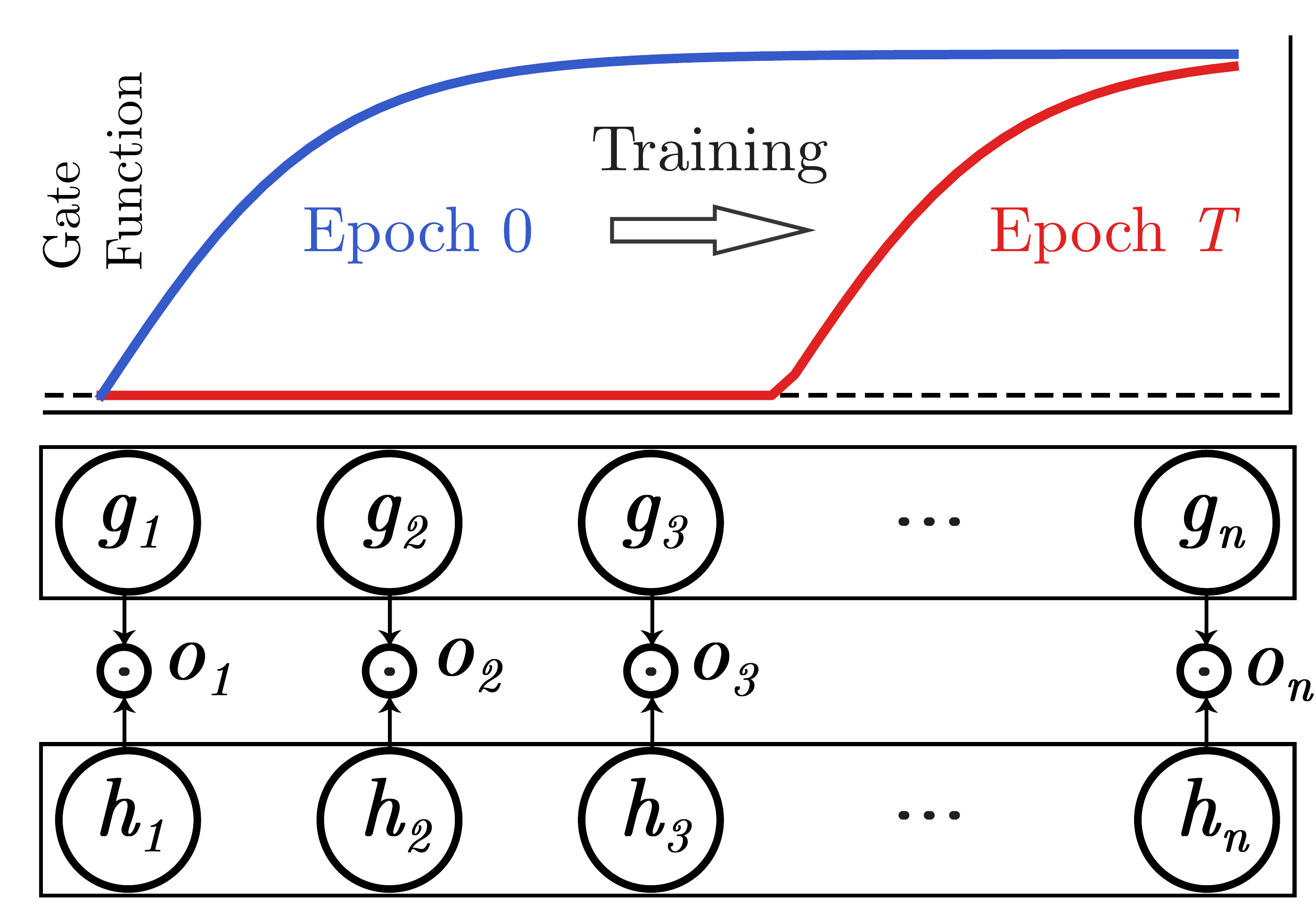}
\captionsetup{font=small}
\caption{Illustration of Discriminative Masking. The gate function shifts to the "right" during training resulting in more zeros in the mask on convergence.}
\label{fig:dam_intro}

\end{wrapfigure}

While there is a growing body of work in structured network pruning \cite{li2016pruning,Liu2017learning,tiwari2021chipnet,gordon2018morphnet}, the basic structure of most state-of-the-art (SOTA) methods in this area (e.g., ChipNet \cite{tiwari2021chipnet} and NetSlim \cite{Liu2017learning}) can be described as training a learnable vector of mask parameters, $\mathbf{g} \in \mathbb{R}^n, \mathbf{g} = [g_1, g_2, ..., g_n]^\intercal$, which when multiplied with the features extracted at a hidden layer, $\mathbf{h} \in \mathbb{R}^n$, results in the pruned outputs of this layer, $\mathbf{o} = \mathbf{g}\circ\mathbf{h}$. Sparsity in the mask parameters is generally enforced using different approximations of the $L_0$ norm of $\mathbf{g}$ (e.g., use of Lasso in NetSlim and use of ``crispness'' loss in ChipNet). Despite recent progress in this area, current SOTA in structured pruning suffer from two key limitations. First, since most methods do not explicitly minimize the $L_0$ norm of $\mathbf{g}$ during pruning, they often suffer from \textit{training instabilities}. In particular, most SOTA methods \cite{Liu2017learning,tiwari2021chipnet} involve thresholding techniques to set small non-zero weights to zero leading to large drops in accuracy during the training process, as evidenced by our results in this paper. 
Second, once the pruning process is complete and we have converged at a compact network, most SOTA methods still need an additional step of \textit{fine-tuning} the network in order to achieve reasonable accuracy. This is not only resource-intensive but there is also an on-going discussion on 
whether and how we should fine-tune \cite{renda2020comparing} or rewind to initial weights \cite{le2021network} or train from scratch \cite{liu2018rethinking}, making it difficult to prefer one approach over another.

At the core of these limitations is the lack of a systematic approach for structured network pruning that jointly prunes and refines weights during training in a single stage, and does not require any fine-tuning upon convergence to achieve SOTA performance. 
Notice that in existing methods for structured pruning, allowing every neuron $j$ to be pruned differently using an independent mask parameter $g_j$ only increases the number of learnable (or free) parameters in the learning objective, increasing the complexity of the problem. Instead, we ask the question: ``{Can we leverage the intrinsic symmetry of neural networks to design a pruning mask using the least number of free parameters?}''



We present a simple solution to this question by proposing a new single-stage structured pruning method that learns compact representations while training and does not require fine-tuning, termed \textit{DiscriminAtive Masking (DAM)}. The basic idea of DAM is to use a monotonically increasing gate function $\mathcal{G}$ for masking every neuron in a layer that only depends on the index $j = 1, 2, ..., n$ (or position) of the neuron in the layer, i.e., $g_j = \mathcal{G}(j)$, and a scalar parameter $\beta$ to be learned during training (see Figure \ref{fig:dam_intro}). At the start of training, the gate function admits non-zero values for all neurons in the layer, allowing all neurons to be unmasked (or active). As the training progresses, the gate function gradually shifts from ``left'' to ``right'' as a result of updating $\beta$ such that upon convergence, only a subset of neurons (on the extreme right) are active while all others are masked out. The key intuition behind DAM is to \textit{discriminatively} prefer some of the neurons (on the right) to be refined (or re-adapted) during the training process for capturing useful features, while gradually masking out (or pruning) neurons on the left. This preferential pruning of neurons using a very simple gate function helps in regulating the number of features transmitted to the next layer.\footnote{This is somewhat analogous to how a physical dam regulates the flow of water by shifting a movable gate.}



\textbf{Contributions:} We show that our proposed DAM approach has remarkably good performance over various applications, including dimensionality reduction, recommendation system, graph representation learning, and structured pruning for image classification, achieving SOTA performance for structured pruning. This shows the versatility of DAM in learning compact representations on diverse problems and network choices, in contrast to baseline methods for structured pruning that are only developed and tested for the problem of image classification \cite{Liu2017learning,tiwari2021chipnet}. Our approach is single-stage, does not require fine-tuning (and hence has lower training time), and does not suffer from training instabilities. We also theoretically show that the learning objective of DAM is directly related to minimizing the $L_0$ norm of the discriminative mask, providing a new differentiable approximation for enforcing $L_0$ sparsity. Our approach is also easy to implement and can be applied to any layer in a network architecture.
The simplicity and effectiveness of DAM provides unique insights into the problem of learning compact representations and its relationship with preferential treatment of neurons for pruning and refining, opening novel avenues of systematic research in this rapidly growing area.

\section{Related Work}

\textbf{Learning with Sparsity:} There is a long history of methods for enforcing compactness (or sparsity) in learned parameters \cite{olshausen1996emergence,tibshirani1996regression}, where it is understood that compact hypotheses can not only result in more efficient data representations but also promote better generalizability \cite{rissanen1986stochastic}. This connection between compactness and generalization has also been recently explored in the context of deep learning \cite{zhou2018compressibility,arora2018stronger}, motivating our problem of learning compact representations.

\textbf{Unstructured Pruning:}
There are several methods that have been developed for unstructured network pruning \cite{louizos2018learning, frankle2018lottery, frankle2020linear, savarese2020winning}, where the goal is to prune individual weights of the network. A widely used technique in this field is referred to as the {Lottery Ticket Hypothesis} (LTH) \cite{frankle2018lottery}, which suggests that a certain subset of weights in a network may be initialized in such a way that they are likely ``winning tickets'', and the goal of network pruning is then to simply uncover such winning tickets from the initialization of an over-parameterized network using magnitude-based pruning. Similar to our work, there exists a line of work in the area of \textit{dynamic sparse training} for unstructured pruning \cite{zhu2017prune, gale2019state,Mocanu2018ScalableTO, Evci2020Rigging, liu2021actually} that gradually prunes the model to the target sparsity during training. However, while unstructured pruning methods can show very large gains in terms of pruning ratio, they are not directly useful for learning compact representations, since the pruned weights, even though sparse, may not be arranged in a fashion conducive to the goal of reducing the number of features.

\textbf{Structured Network Pruning:}
The goal of structured pruning is to enforce sparsity over the network substructures (e.g., neurons or convolutional channels) that is directly relevant to the goal of learning compact representations. 
Most structured pruning SOTA methods need a three-stage process \cite{NIPS2015_ae0eb3ee}, i.e., training, pruning and finetuning, in order to get highly compressed models. Early exploration on the $L_1$ norm based filter pruning can be traced back to \cite{li2016pruning}, followed by \textit{Network Slimming} (NetSlim) \cite{Liu2017learning}, which uses a sparsity-inducing regularization based on $L_1$ norm. Following this line of work, a recent development in the area of network pruning is \textit{ChipNet} \cite{tiwari2021chipnet}, which introduces an additional \textit{crispness loss} to the $L_1$ regularization, and achieves SOTA performance on benchmark tasks. However, network slimming can only work if the underlying architecture has batch normalization layers. It also has training instabilities due to soft-thresholding inherent to Lasso. ChipNet also iterates between soft- and hard-pruning, which is done after the training phase. Further, both these approaches require finetuning or retraining, which can be resource intensive.

Faster pruning schemes have also been explored by previous works. Bai et al \cite{bai2020few} proposed a few-shot structured pruning algorithm utilizing filter cross distillation. The SNIP \cite{lee2018snip} took one step ahead by performing single-stage pruning. However, SNIP falls in the category of unstructured pruning. To the best of our knowledge, no existing structured pruning method has been demonstrated to achieve SOTA performance without finetuning. For example, while the method proposed in \cite{louizos2018learning} can be potentially used as a single-stage structured pruning method with the help of group sparsity, the empirical analysis of the paper only focused on unstructured pruning.

\textbf{Deep Representation Learning}:
Leveraging the power of deep neural networks, high-dimensional data can be encoded into a low-dimensional representation \cite{Hinton504}. Such process is known as deep representation learning. A good representation extracts useful information by capturing the underlying explanatory factors of the observed input. The existence of noises, spurious patterns and complicated correlations among features make it a challenging problem. Previous work \cite{autoencoder} show that an Autoencoder architecture can be used for denoising images. To disentangle correlated features, VAE \cite{kingma2013auto} and $\beta$-VAE \cite{higgins2016beta} encourages the posterior distribution over the generative factors $q(z|x)$ to be closer to the isotropic Gaussian $\mathcal{N}(0, I)$ that explicitly promotes disentanglement of the latent generative factors. Driven by the idea of the autoencoder, the deep representation learning has been found remarkably successful in various of applications, e.g., graph representation learning \cite{kipf2016variational, davidson2018hyperspherical, di2020mutual, mavromatis2020graph, yang2018binarized, pan2018adversarially} and recommendation system \cite{zhang2019inductive}. 

However, most existing works in representation learning treat the embedding dimension as a hyperparameter, which can be crucial and difficult to choose properly. Small embedding dimensions cannot sufficiently represent the important information in the data that leads to bad representations. On the other hand, large embedding dimensions allow some level of redundancy in the learned representations, even picking up spurious patterns that can degrade model performances.

\section{Proposed Approach}
\label{sec:method}

\begin{figure}[t]
    \centering
    \includegraphics[width=\linewidth]{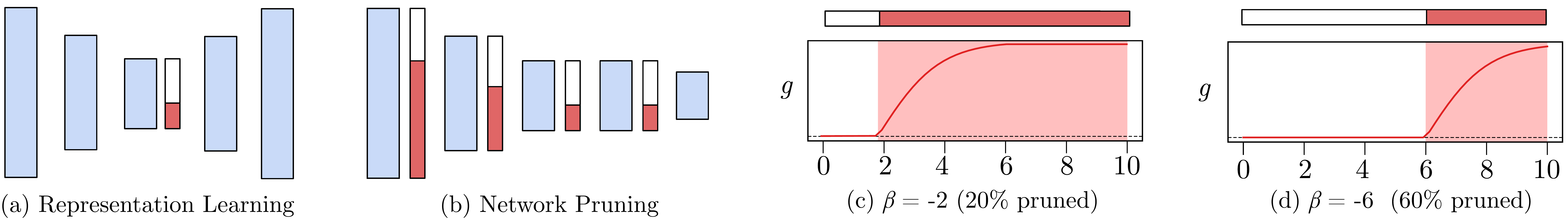}
    \caption{Illustration of the problems of representation learning (a) and structured network pruning (b). Blue blocks are layers in neural networks while red blocks show non-zero values in the mask layers. 
    Figure (c) and (d) show two states of the gate function for $k=10$. As an example, $\beta$=-2 in (c) prunes 20\% of the neurons while $\beta$=-6 in (d) prunes 60\% of the neurons.}
    \label{fig:dam}
    
\end{figure}

\textbf{Problem Statement:}
Let us represent the general architecture of a neural network with $l$ layers as $\mathcal{F}(\bm{x}) = \Psi_l \Psi_{l-1} ... \Psi_2 \Psi_1 \bm{x}$, where $\bm{x}$ are the input features to the network and $\bm{h}_i = \Psi_i(\bm{h}_{i-1})$ are the learned features at layer $i$. 
Let us also consider the ``masked'' version of this network where the features learned at every hidden layer $i$ are multiplied with a continuous mask layer $G_i$ in the following manner, $\mathcal{F}_{\bm{g}}(\bm{x}) = \Psi_l G_{l-1} \Psi_{l-1} ... G_2 \Psi_2 G_1 \Psi_1 \bm{x}$, where $\bm{g}_i = G_i(\bm{h}_{i})$. The goal of learning compact representations can then be framed as minimizing the following learning objective:
\begin{align}
    \min_{\theta}\, \mathbb{E}_{\bm{x} \sim \mathbb{D}_{tr} } \left[ \mathcal{L}\left( \mathcal{F}_{\bm{g}}(\bm{x}) \right) + \frac{\lambda}{l-1} \sum_{i=1}^{l-1}\norm{\bm{g}_i}_0 \right],
\end{align}
where $\theta$ are the learnable parameters of the network\footnote{Note the parameter set $\theta$ includes the learnable parameter of DAM layers $\beta$.}, $\mathbb{D}_{tr}$ is the training set, $\mathcal{L}(.)$ is the loss function, and $\lambda$ is the trade-off parameter for enforcing $L_0$ sparsity in the masked outputs, $\bm{g}_i$. Note that this formulation provides a unified view of both representation learning and structured network pruning, which differ in the choice of loss function (typically, reconstruction error for representation learning and cross-entropy for structured pruning), and the number of layers that are pruned (in representation learning, we only aim to prune the bottleneck layer while in structured pruning, we aim to prune all of the layers in the network, see Figure \ref{fig:dam}). Further, notice that we only enforce $L_0$ penalty on the gate outputs $\bm{g}_i$, while the hidden features $\bm{h}_i$ are allowed to be non-sparse.

\textbf{DiscriminAtive Masking (DAM)}: 
To describe the key idea of our proposed DAM approach, we first introduce the concept of an `ordering' among the neurons at any layer $i$, such that we \textit{discriminatively} (or preferentially) focus on pruning neurons lower in the order, while retaining and refining the features learned at neurons higher in the order. Specifically, let us assign every neuron $j$ at layer $i$ with an ``order number,'' $\mu_{ij}$. The masking operation of DAM can then be implemented using a simple gate function $\bm{g}_i$, whose value at neuron $j$ monotonically increases with the order number of the neuron, $\mu_{ij}$, as follows:
\begin{align}
    g_{ij} = \text{ReLU}\left[ \tanh\left( \alpha_i (\mu_{ij} + \beta_i) \right) \right] = \max{ \left[ \tanh\left( \alpha_i (\mu_{ij} + \beta_i) \right), 0 \right]}, \label{eq:gate_function}
\end{align}
where $\alpha_i$ is a constant scalar parameter (termed as the \textit{steepness} parameter) that is not optimized during training, while $\beta_i$ is a `learnable' scalar parameter (termed the \textit{offset} parameter) that is the \textit{only} parameter optimized during training to control (or regulate) the process of masking in DAM. Figures \ref{fig:dam}(c) and (d) show examples of this gate function (in red) at two different values of $\beta_i$. Note that using a single learnable parameter $\beta_i$ to implement the masking process introduces significant simplicity in the design of the gate function, in contrast to SOTA methods in structured pruning \cite{Liu2017learning,tiwari2021chipnet} where a different masking parameter is trained for every neuron in the network. We hypothesize this simplicity of the gate function, in conjunction with the preferential pruning property of DAM, to result in remarkable gains in learning compact representations, as evidenced later in our experiments. Further, note that while we can use any monotonically increasing function to implement our gates, we found ReLU-tanh to perform best in practice.




\textbf{Neuron Ordering in DAM}: 
We now discuss our approach for assigning neuron order numbers $\mu_{ij}$ for DAM to work. Note that the order numbers only have to be assigned during the initialization step, and do not change during the training process. Further, note that every neuron is initialized with weights that are independent and identically distributed (because of the i.i.d nature of commonly used initialization methods in deep learning \cite{sutskever2013importance}). Hence, all neurons are symmetrically invariant to each other at initialization and any neuron has an equally good chance of refining themselves to capture useful features later in the training process as any other neuron. Hence, any random ordering of the neurons in the DAM approach would result in a similar pruning process (we empirically demonstrate the permutation invariance of $\mu_{ij}$ in Section 5). For simplicity, we consider a trivial choice of $\mu_{ij} = kj/n_i$, where $k$ is a constant parameter that determines the domain size of $\mu_{ij}$, and $n_i$ is the total number of neurons at layer $i$.

\textbf{How Does Reducing $\beta_i$ Enforce $L_0$ Sparsity?}
Now that we have described the implementation of our gate function $\bm{g}_i$, let us understand how $\bm{g}_i$ behaves as we change the only learnable parameter, $\beta_i$. Figure \ref{fig:dam} (c) shows an example of $g_i$ when $\beta_i = -2$. At this stage, we can see that only a small number of neurons have zero gate values (white). As we reduce $\beta_i$ to $-6$ in Figure \ref{fig:dam} (d), we can see that the gate moves towards the 'right', resulting in more excessive pruning of neurons. In general, the number of non-zero values of the gate function (and hence, its $L_0$ norm) is directly related to $\beta_i$ as follows:
\begin{equation}
\label{eq:l0_norm}
    \norm{\bm{g}_i}_0 = \sum_{j=1}^{n_i} \mathbb{1}(g_{ij}>0) = \sum_{j=1}^{n_i} (1 - \mathbb{1}(\mu_{ij} \leq \beta_i)) = \ceil{n_i(1 + \beta_i /k)},\quad \text{for} \quad \beta_i \geq -k,
\end{equation}
where $\mathbb{1}$ indicates the identity operator and $\ceil{\cdot}$ is the ceiling operator. In order to back-propagate gradients, we use a continuous approximation of Equation (\ref{eq:l0_norm}) (by dropping the ceiling operator) for regularization as $\norm{\bm{g}_i}_0 \approx n_i(1 + \beta_i /k)$.
Hence, minimizing $L_0$ norm of $\bm{g}_i$ is directly proportional to minimizing $\beta_i$ with a scaling factor of $n_i/k$.

\textbf{Learning Objective of DAM:}
While minimizing $\beta_i$ with scaling factors proportional to $n_i$ can explicitly minimize $L_0$ norm of the gate values, we found in practice that layers with smaller number of neurons $n_i$ still require adequate pruning of features for learning compact representations proportional to layers with large $n_i$. Hence, in our learning objective, we drop the scaling factor and directly minimize the sum of $\beta_i$ across all layers in the following objective function.
\begin{align}
    \label{eq:network_pruning_rewrite}
    \min_{\theta}\, \mathbb{E}_{x \sim \mathbb{D}_{tr} } \left[ \mathcal{L}\left( \mathcal{F}_g(\bm{x}) \right) + \frac{\lambda}{l-1} \sum_{i=1}^{l-1} \beta_i \right],
\end{align}



\textbf{Theoretical Analysis of DAM:} Analysis of the dynamics of the gate functions and additional specifications for choosing $\alpha_i$ and the gate function are provided in Appendix A.

\textbf{Parameter Specification and Implementation Details:}
In all of our experiments, we used $\alpha_i=1$ and $k=5$. In our structured network pruning experiments, we used a cold-start of 20 epochs (i.e., the $\beta_i$'s were frozen for the duration of cold-start), so as to allow the leftmost neurons to undergo some epochs of refinement before beginning the pruning process. We also set the initial value of $\beta_i$ to 1, which can also be thought of as another form of cold-starting (since pruning of a layer only starts when $\beta_i$ becomes less than zero).

 
\section{Experiments on Representation Learning Problems}
\label{sec:CRL}

\subsection{Results on Dimensionality Reduction (DR) Problems}

\textbf{Problem Setup:}
We evaluate the effectiveness of DAM in recovering the embeddings of synthetically generated dimensionality reduction problems. The general form of the problem is expressed by (\ref{eq:rd_Dam_objective}).
Suppose $\Omega \in \mathbb{R}^r$ and all of its $r$ dimensions are independent from each other. A $d$-dimensional data $X$ can be expressed using a transformation $\Psi: \mathbb{R}^r \longrightarrow \mathbb{R}^d$ defined on Hilbert space, and $X = \Psi {\Omega}$, $d > r$.
Let $G = \diag{(\bm{g})}$, we formulated the dimensionality reduction problem as:
\begin{align}
    \min_{\Theta}\left\{ \norm{{\mathcal{F}}_D \cdot G \cdot \mathcal{F}_E \cdot X - X}_F^2 + \lambda \beta \right\} ,\quad \Theta = \{ \theta_E, \theta_D, \beta \}
    \label{eq:rd_Dam_objective}
\end{align}
The $\mathcal{F}_E, G, \mathcal{F}_D$ are trained end-to-end using gradient descent. In our experiments, $\Omega$ is sampled from a isotropic normal distribution, i.e., $\Omega \sim{ \mathcal{N}_r(\bm{0}, I_{r})}$. We let $\Psi$ to use the same structure as the decoder $\mathcal{F}_D$ with randomly generated parameters $\theta_\Psi$, i.e., $\Psi = \mathcal{F}_D|_{\theta_D=\theta_\Psi}$, so that the minimum number of dimensions needed in the encoded representation for the decoder to reconstruct $X$ is $r$. 

\textit{Linear DR:} 
We test DAM for removing linear correlations in a given rank-deficient matrix. In this case, $\Psi$ is a matrix representing a linear projection from $\mathbb{R}^r$ to $\mathbb{R}^d$. We use two real matrices as the encoder and decoder and train the DAM to find the full-rank representation of $X$. 

\textit{Nonlinear DR:} 
To test the ability of DAM for disentangling nonlinearly correlated dimensions, we present two cases: (i) $\Psi$ is a polynomial kernel of degree 2 (we use a QRes layer from \cite{jie2021qres}); (ii) $\Psi$ is a nonlinear transformation expressed by a neural network. We use a deep neural network as the encoder with sufficient complexity.




\begin{figure}[ht]

\begin{subfigure}{.33\textwidth}
  \centering
  \includegraphics[width=.9\linewidth]{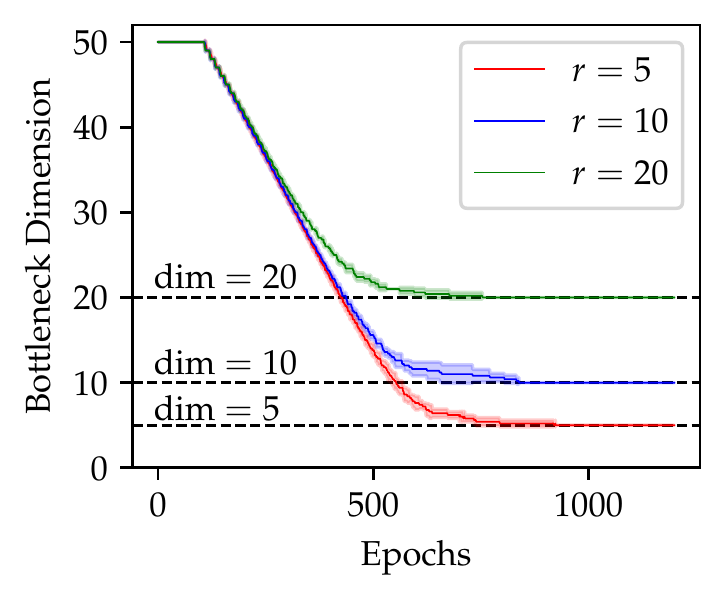}
  
  \caption{Linear $\Psi$}
  \label{fig:linear_rd}
\end{subfigure}
\begin{subfigure}{.33\textwidth}
  \centering
  \includegraphics[width=.9\linewidth]{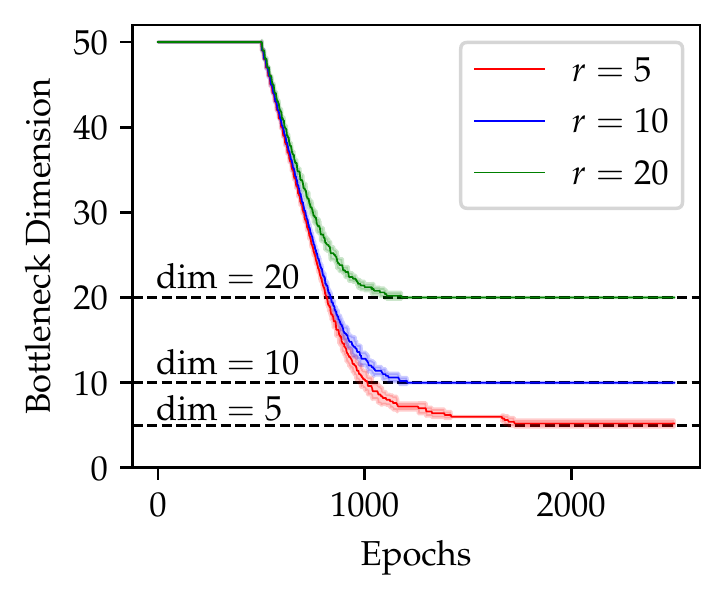}
  
  \caption{Polynomial $\Psi$}
  \label{fig:quad_rd}
\end{subfigure}
\begin{subfigure}{.33\textwidth}
  \centering
  \includegraphics[width=.9\linewidth]{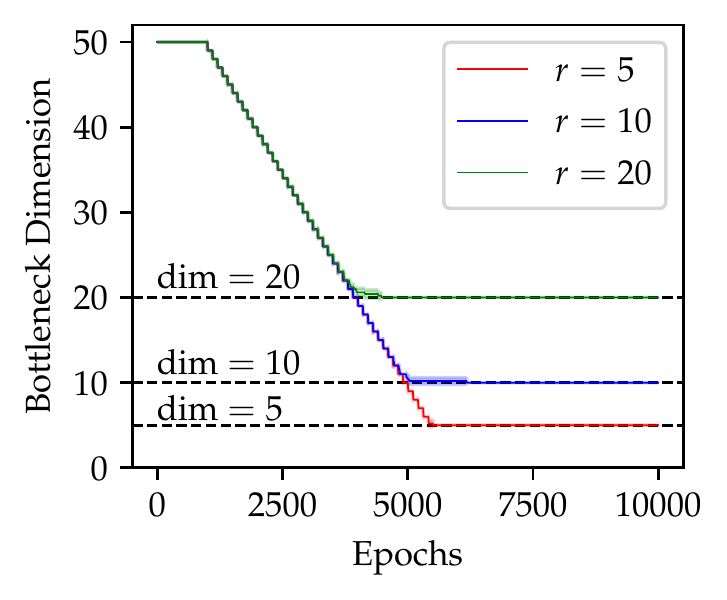}
  
  \caption{Neural Network $\Psi$}
  \label{fig:mlp_rd}
\end{subfigure}

\caption{
Bottleneck dimension (of nonzero entries) vs number of epochs during training.
}
\label{fig:dimensionality reduction}

\end{figure}

\textbf{Observation:}
Figure \ref{fig:dimensionality reduction} shows the convergence of bottleneck dimensions over the training epochs for different synthetic data with varying sizes of underlying factors ($r$). The curves and shades show the mean and standard deviation of five different runs. We can see that DAM consistently uncovers the exact dimension of the underlying factors $r$. We provides details of in-depth sensitivity analysis of DAM results based on its hyper-parameters (e.g., learning rate and $\lambda$) in Appendix C. 

\textbf{Theoretical Analysis of DAM for linear DR:}
We theoretically show that DAM  converges to the optimal solution for the linear DR case (see Appendix B for details).



\subsection{Recommendation System Results} 

\textbf{Problem Setup:}
We consider the state-of-the-art method for recommendation system problems (IGMC \cite{zhang2019inductive}), which transforms the rating matrix completion task to a link prediction problem, where the ratings are interpreted as links between users and items. 


 In particular, IGMC generates 128-dimensional embeddings to represent the enclosing subgraph and further avoids overfitting using a dropout layer with a dropout rate of 0.5. We replaced the dropout layer with a DAM layer in the IGMC to reduce the dimension of the learned representations. We train our IGMC-DAM model for 100 epochs under the same training configurations.

\begin{table}[ht]
\small
\centering
\caption{Results of IGMC with and w.o. DAM on recommendation system tasks.\\}
\begin{tabular}{ccccc}
\toprule
            & \multicolumn{2}{c}{IGMC}          & \multicolumn{2}{c}{IGMC-DAM}                       \\ 
            & RMSE                  & Dimension & RMSE                           & Dimension         \\ \hline
Flixter     & $0.8715 \pm 0.0005$   & 128       & $\mathbf{0.8706 \pm 0.0003}$   & $\mathbf{32}$     \\ 
Douban      & $0.7189 \pm 0.0002$   & 128       & $\mathbf{0.7183 \pm 0.0004}$   & $\mathbf{17}$     \\ 
Yahoo-Music & $19.2488 \pm 0.0123$  & 128       & $\mathbf{19.0166 \pm 0.0444}$  & $\mathbf{83}$     \\ 
\bottomrule
\end{tabular}
\label{tab:recommender}
\end{table}

 \textbf{Observation:}
 Table \ref{tab:recommender} shows that DAM successfully reduces the dimensions of the learned representations without any increase in errors, demonstrating that DAM is able to boost the performance of IGMC by learning compact representations over the target graph.

\subsection{Graph Representation Learning Results} 

\textbf{Problem Setup:}
 Following the previous experiment, we further explore the effectiveness of DAM in learning compact graph representations. Our goal is to learn low-dimensional embeddings for each node in a graph that captures the structure of interaction among nodes. A simple graph autoencoder, e.g., GAE \cite{kipf2016variational}, uses a graph convolutional network (GCN) \cite{kipf2016semi} as an encoder and an inner product as a decoder. The encoder calculates the embedding matrix $Z$ from the node feature matrix $X$ with the adjacency matrix $A$, and the decoder reconstructs a adjacency matrix $\hat{A}$ such that: $\hat{A} = \sigma(ZZ^\top)$, with $Z = \text{GCN}(X, A)$. The reconstruction loss, BCE$(A, \hat{A})$ is backpropagated to train the GAE model. To reduce the dimension of the learned representation, we add a DAM layer in a GAE after the encoder (GAE-DAM). 

\begin{figure}[ht]
\begin{subfigure}{.33\textwidth}
  \centering
  \includegraphics[width=.9\linewidth]{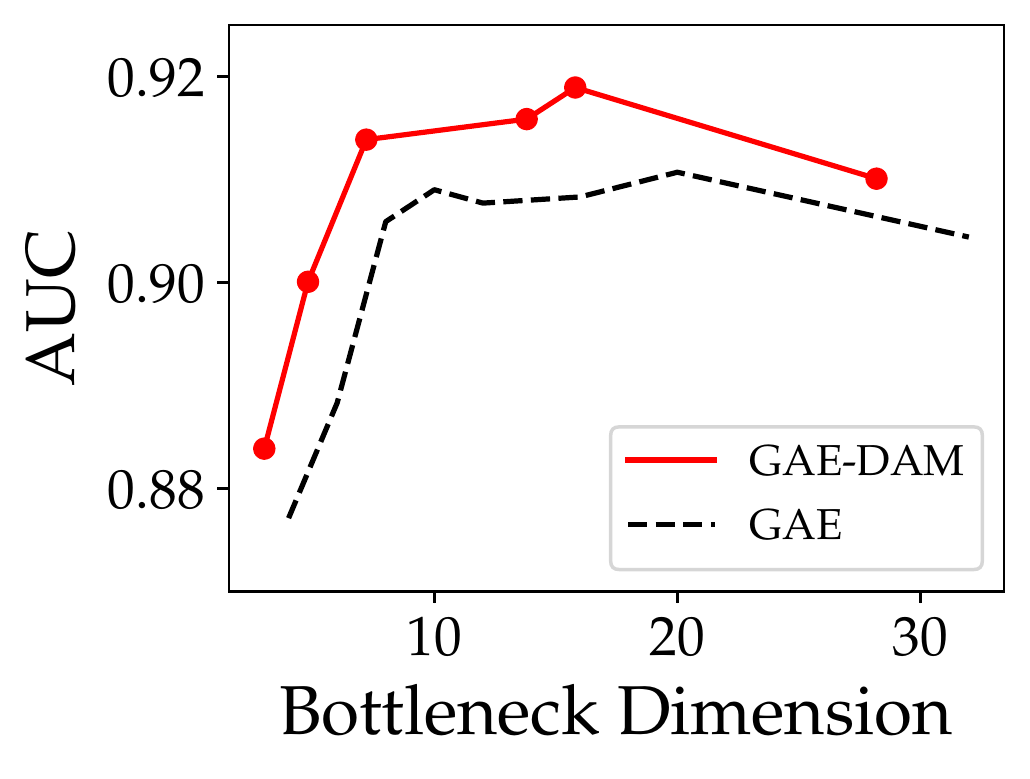}
  
  \caption{Cora}
  \label{fig:cora_lp}
\end{subfigure}
\begin{subfigure}{.33\textwidth}
  \centering
  \includegraphics[width=.9\linewidth]{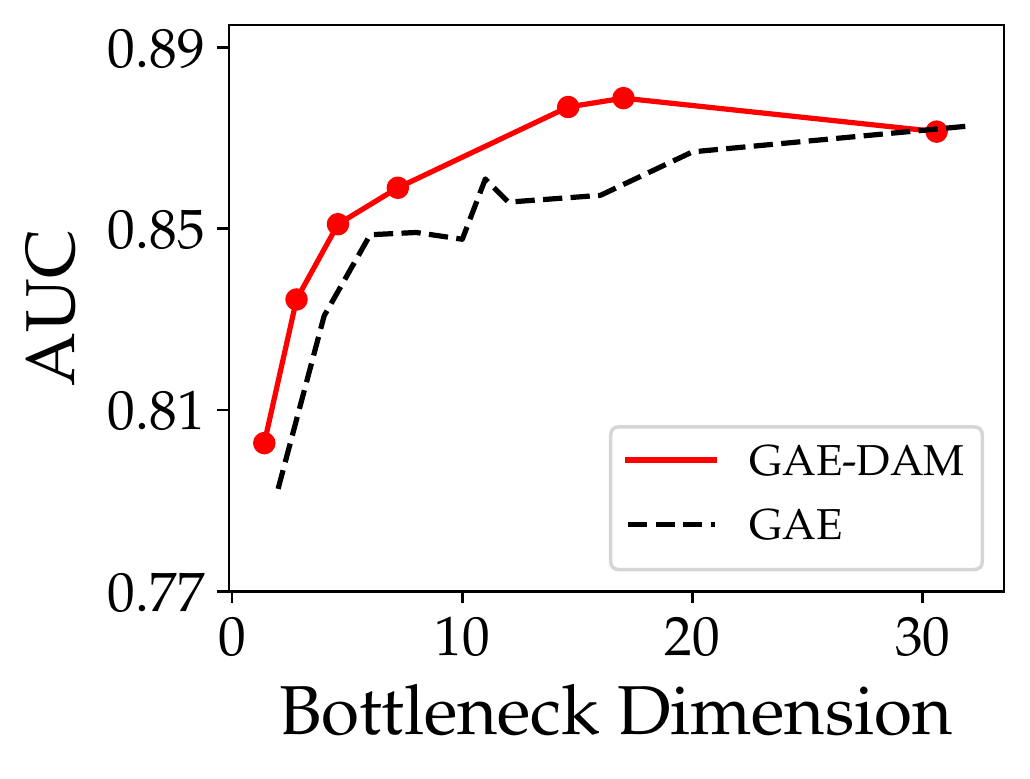}
  
  \caption{CiteSeer}
  \label{fig:citeseer_lp}
\end{subfigure}
\begin{subfigure}{.33\textwidth}
  \centering
  \includegraphics[width=.9\linewidth]{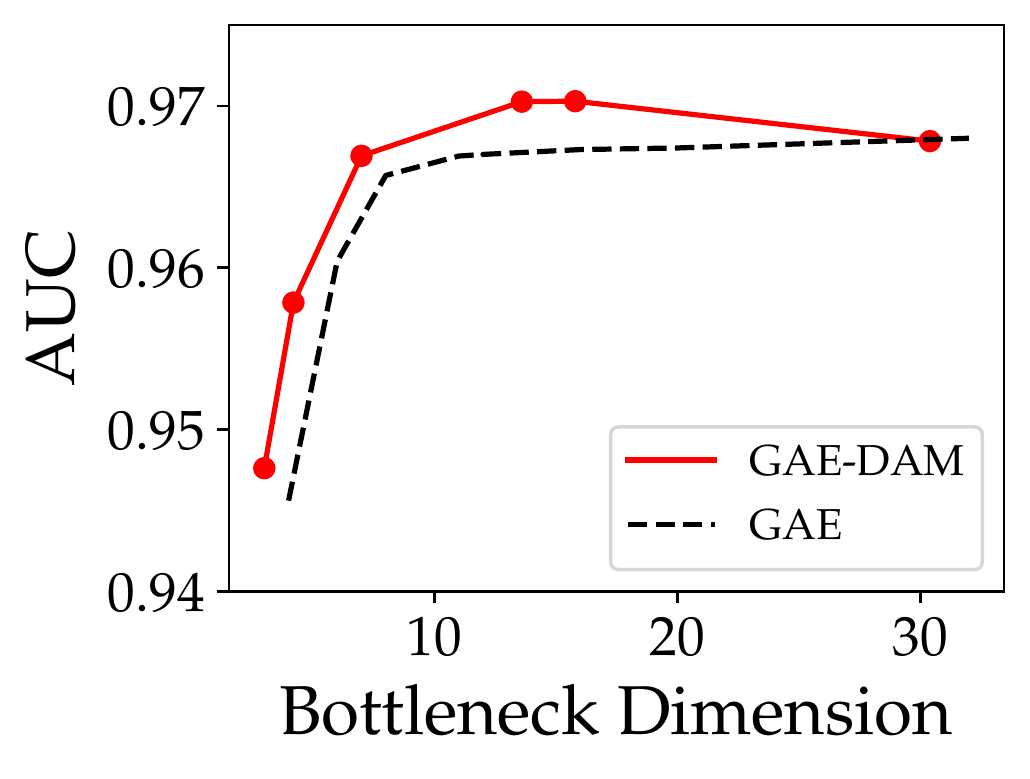}
  
  \caption{PubMed}
  \label{fig:pubmed_lp}
\end{subfigure}

\caption{
Link prediction performance for Cora, CiteSeer and PubMed Dataset. Competing models are GAE, GAE-DAM.}
\label{fig:link_prediction}
\end{figure}

 \textbf{Observation:}
Figure \ref{fig:link_prediction} shows that DAM based GAE method is able to learn meaningful structural information in compact latent embeddings. DAM improves the link prediction performance of the simple GAE for Cora, CiteSeer and PubMed dataset by learning compact representations.

\subsection{Representation Learning Results for MNIST Dataset} 

\textbf{Problem Setup:}
We further  evaluate DAM on the problem of dimensionality reduction using simple auto-encoders on the MNIST dataset, and also compare the effectiveness of DAM (that directly enforces $L_0$ sparsity) with an ablation of our approach that instead minimizes the $L_1$ norm (similar to Lasso). We vary the trade-off parameter $\lambda$ to compare the sparsity of the learned representations between DAM and $L_1$-norm based ablation method.

\begin{figure}[ht]
\begin{subfigure}{.33\textwidth}
  \label{fig:rec-autoenc}
  \centering
  \includegraphics[width=.99\linewidth]{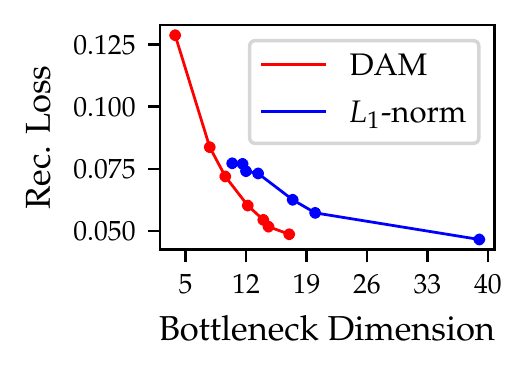}
\end{subfigure}
\begin{subfigure}{.33\textwidth}
  \label{fig:f1-autoenc}
  \centering
  \includegraphics[width=.99\linewidth]{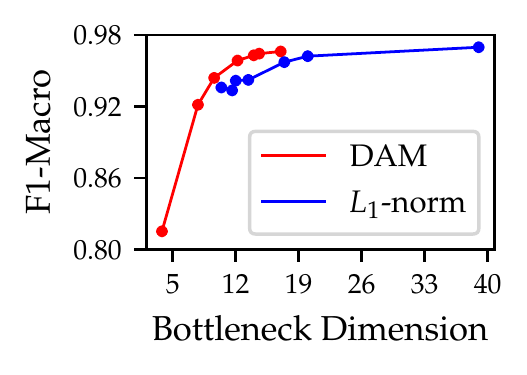}
\end{subfigure}
\begin{subfigure}{.33\textwidth}
  \label{fig:lam-btl-autoenc}
  \centering
  \includegraphics[width=.99\linewidth]{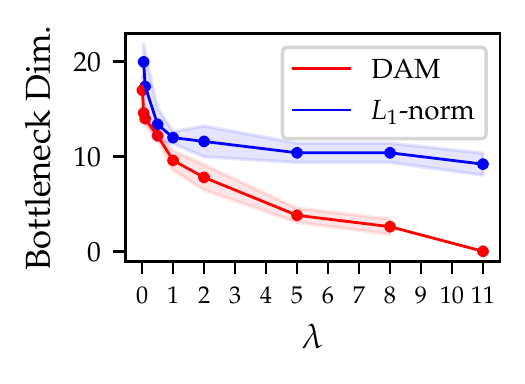}
\end{subfigure}

\caption{
Performance comparison plots of DAM with $L_1$-norm method on MNIST dataset.
}
\label{fig:aenc_l1_dam}
\end{figure}

\textbf{Observation:}
Figure \ref{fig:aenc_l1_dam} shows that DAM is able to achieve lower reconstruction error as well as higher $F1$ scores over the same bottleneck dimensions than $L_1$ based method. 
In Figure \ref{fig:aenc_l1_dam} (c), we can see that DAM shows a near-linear descending trend of bottleneck dimensions as $\lambda$ increases (further evidence presented in Appendix E1). In contrast, the $L_1$ based method has a saturating effect at large $\lambda$ values since the weights come close to zero but require some thresholding to be pruned. This shows that DAM is amenable to learning highly compact representations in contrast to $L_1$ based methods. 


\section{Experiments on Structured Network Pruning}

\textbf{Evaluation Setup}
Here we demonstrate the effectiveness of DAM on the task of structured network pruning. We compared the performance of our proposed DAM with two SOTA structured pruning algorithms, Network Slimming (Net-Slim) \cite{Liu2017learning} and ChipNet \cite{tiwari2021chipnet}. Net-Slim is well-established SOTA that is widely used as a strong baseline, while ChipNet is the latest SOTA representing a very recent development in the field. ChipNet and Net-Slim are both pretrained for 200 epochs, while the DAM performs single-stage training and pruning in the same number of epochs. Since, our DAM approach does not require additional fine-tuning, we imposed a limited budget of 50 epochs for fine-tuning the SOTA methods. Note that ChipNet also has a pruning stage which involves 20 additional epochs. We evaluate the performance of these methods on the PreResnet-164 architecture on benchmark computer vision datasets, CIFAR-10 and CIFAR-100. Additional evaluation setup details are in Appendix E2.

\begin{figure}[ht]
\begin{subfigure}{.32\textwidth}
  \centering
  \includegraphics[width=.99\linewidth]{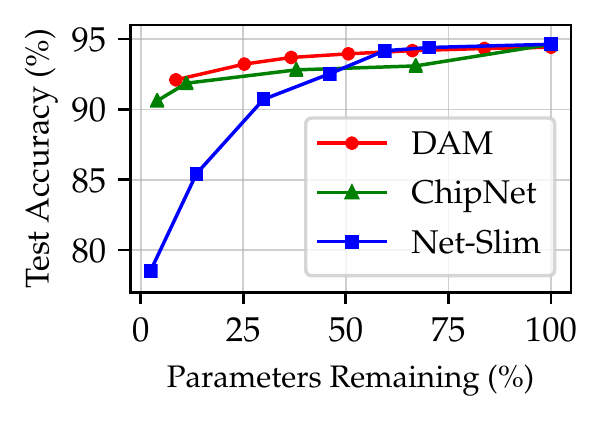}
  
  \caption{CIFAR-10}
  \label{fig:c10}
\end{subfigure}
\begin{subfigure}{.32\textwidth}
  \centering
  \includegraphics[width=.99\linewidth]{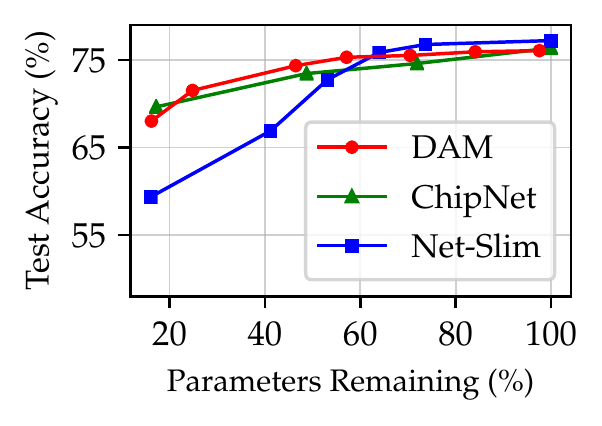}
  
  \caption{CIFAR-100}
  \label{fig:c100}
\end{subfigure}
\begin{subfigure}{.34\textwidth}
  \centering
  \includegraphics[width=.99\linewidth]{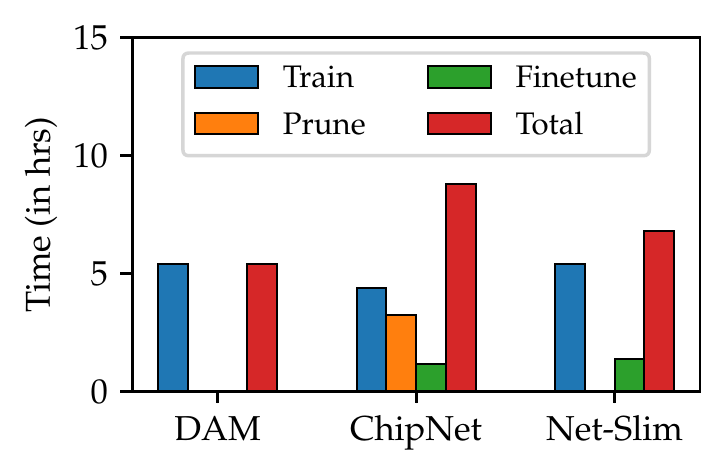}
  
  \caption{Training Time}
  \label{fig:c10_time}
\end{subfigure}

\caption{
Performance and running time comparison of DAM with state-of-the-art structured pruning methods on PreResNet-164 for various parameter pruning ratios.
}
\label{fig:cifar}

\end{figure}

\textbf{Performance and Running time comparison}: Figures \ref{fig:c10} and \ref{fig:c100} demonstrate the performance of different network pruning methods for various pruning ratios. We observe that for both datasets, DAM is able to outperform Net-Slim by a significant margin specially when the models are sparse. Additionally, DAM is able to achieve similar or slightly better performance than ChipNet with no additional fine-tuning. We also compare the total running time for pruning across the different models (divided into the three categories) in Figure \ref{fig:c10_time}. We can see that the training time of the three methods are almost the same, with ChipNet being slightly small. 
However, the running time for the pruning and fine-tuning stages for DAM are both zero. Net-Slim also does not involve additional pruning after training. However, ChipNet involves 20 epochs of pruning which is significant and is almost comparable to its pretraining time. Finally, comparing the total running time taken by each of the structured pruning methods, we observe that DAM is significantly faster than the current SOTA counterparts owing to its single-stage pruning approach.   

\textbf{Stability Analysis}: We further analyse the stability of our proposed DAM approach for network pruning through visualization of the training dynamics. We observe that the training cross-entropy (CE) loss and the validation CE loss are very similar to what we expect from a Preresnet model trained on benchmark vision datasets with learning rate changes at 100 and 150 epochs, respectively. We further notice that the total training loss for the DAM is also very stable and does not show any sudden variations, leading to steady convergence. Finally, from the convergence plot for $\mathbb{E}[\beta_i]$ (green-dashed line), we can see that our pruning does not involve any sudden steps and happens in a continuous and steady rate throughout the training stage. 

The training dynamics of the ChipNet (blue shaded region in the middle panel) is exactly the same as training vanilla vision models since it does not involve sparsity constraints. However, as we move to the pruning stage, we notice a sharp rise in both training and validation losses (white region). This is due to the use of a different optimizer AdamW as opposed to SGD which was used in the training stage. Further, a very interesting phenomenon happens in the fine-tuning stage (red shaded region) where the training and validation losses increases for the first 25 epochs and then takes a sharp decrease once the learning rate is adjusted. This suggests that after the pruning stage, finetuning the model with large learning rate forces the model to make large updates in the initial epochs (akin to training from scratch), which further drives the model out of the local minima. Once the learning rate is adjusted, the model is able to slowly progress towards a new local optima and ultimately converges to the fine-tuned minima.  

Finally, both the training and fine-tuning dynamics of the Net-Slim (rightmost panel) appears to be highly unstable. This suggests that the pre-training with sparsity loss is not stable and can lead to large fluctuations in the training loss. Also, note that Net-Slim does not have a validation set as it instead uses the test set to choose the best model. We have refrained from using the best model by looking at the performance on the test set, and instead evaluate the model from the last epoch. 

\begin{figure}[ht]
    \label{fig:stability}
    \centering
    \includegraphics[width=\textwidth]{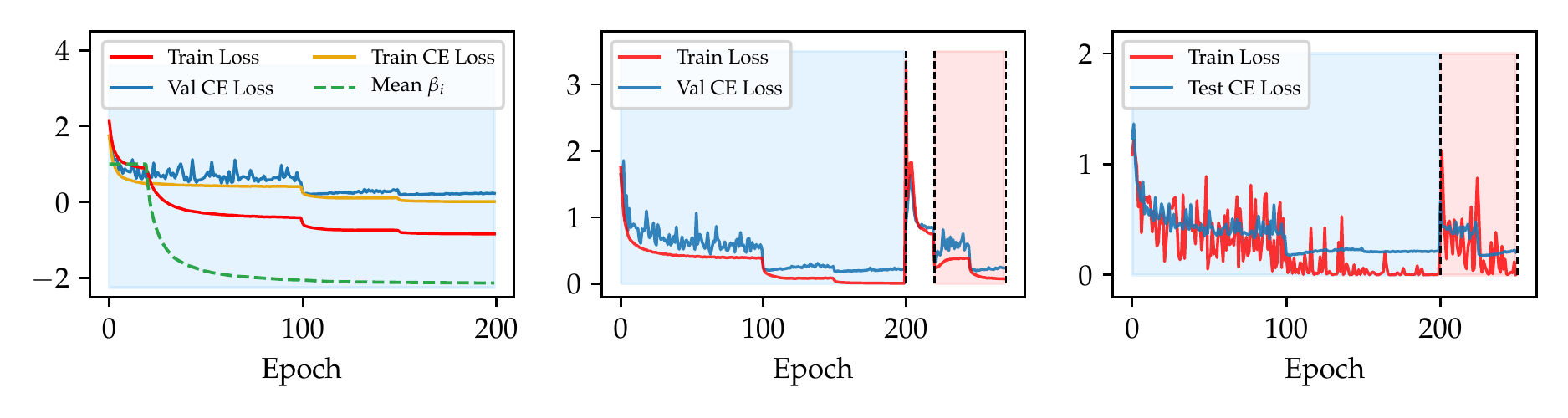}
    \caption{Stability analysis for DAM (left), ChipNet (middle) and Network Slimming (right) through visualization of the loss curves. Blue, white and red shades denote the training, pruning and fine-tuning stages respectively. }
\end{figure}

\textbf{Permutation Invariance}: To test the effect of permuting the neuron order on the performance of DAM, Table \ref{tab:permutation_invariance} provides results across five randomly permuted values of $\mu_{ij}$ across every neuron in the network, while using the same initialization as previous experiments. We can see that DAM is invariant to permutations of neuron order during initialization, validating our simple choice of assigning neuron ordering based on their index values. 

\begin{table}[ht]
\small
\centering
\caption{Permutation Invariance results on PreResNet-164 on CIFAR-10. RSD denotes the relative standard deviation.\\}
\label{tab:permutation_invariance}
\begin{tabular}{cclllcc}
\thickhline
 & Run 1 & Run 2 & Run 3 & Run 4 & Run 5 & RSD \\ \hline
Test Accuracy  & 94.11\% & 93.86\% & 93.71\% & 93.93\% & 93.52\% & 0.0024 \\ 
Channels Pruned & 43.31\% & 43.14\% & 43.31\% & 43.14\% & 43.02\% & 0.0029 \\ 
Parameters Pruned & 64.01\% & 63.40\% & 63.97\% & 63.64\% & 63.37\% & 0.0048 \\ \thickhline
\end{tabular}

\end{table}

\textbf{Additional Results:}
To illustrate generalizibility we provide additional network pruning results for more networks such as VGG-19, PreResnet-20, PreResNet-56 and PreResnet-110 on CIFAR datasets and LeNet-5 on MNIST in Appendix E3. To demonstrate decorrelated kernels obtained after pruning using DAMs we use CKA similarity \cite{kornblith2019similarity}. The results can be found in Appendix E3. For extreme values of $\lambda$, we demonstrate that DAM is able to pruning entire bottleneck layers for ResNet architectures by visualizing the pruned architectures in Appendix E4. 

\section{Discussion, Open Questions, and Future Works}
\label{sec:conclusion}
\vspace{-1ex}
The effectiveness of DAM in learning compact representations provides a novel perspective into structured network pruning and raises several open questions about its relationships with other concepts in this area. In the following, we list some of these connections and describe the limitations and future extensions of our work.

\textbf{Connections with Lottery Ticket Hypothesis}:
Our work builds on the intuition of discriminative masking, i.e., we preferentially favor some neurons to be refined while favoring some other neurons to be pruned. This has interesting connections with the lottery ticket hypothesis (LTH) \cite{frankle2018lottery}, where some subsets of initialized weights in the network are hypothesized to contain ``winning tickets'' that are uncovered after a few epochs of training. While there is some open discussion on whether the winning tickets exist even before initialization \cite{frankle2019stabilizing, kathrin2021winingticket}, there are several studies supporting this hypothesis \cite{savarese2020winning, frankle2020linear}. 
Our results attempts to throw light on the question: ``can we find winning tickets if we discriminatively search for it in a certain subregion of the network?'' Further, we are able to show that the results of DAM are invariant to random permutation of the neuron indices at initialization, since all neurons receive identically distributed weights. Along these lines, we can also explore if there are certain ordering of neurons (e.g., in accordance with their likelihood of containing winning tickets revealed through LTH) that can perform better than random ordering in DAM.

\textbf{Connections with Dropout:} Another interesting connection of DAM is with the popular regularization technique of Dropout \cite{srivastava2014dropout} that  randomly drops out some neurons during training to break ``co-adaptation'' patterns in the learned features so as to avoid overfitting. Essentially, by randomly dropping a neuron at some epoch of training, the other neurons are forced to pick up the learned features of the dropped neuron, thus resulting in the learning of robust features. This is similar to the ``re-adaptation'' of weights at the active neurons at some epoch of DAM training, while the mask value of the neurons in the transitioning zone (when $0 < g_j < 1$) are gradually dropped by the shifting of the gate function. This motivates us to postulate the \textit{weight re-adaptation hypothesis} as a potential reason behind the effectiveness of DAM in learning compact representations, which requires further theoretical justifications.


\textbf{Budget-aware Variants of DAM}:
One promising future extension of our current DAM formulation would be to make it budget-aware,  i.e., to stop the training process once we arrive at a target level of sparsity. Note that 
there is a direct correspondence between the value of the learnable gate offset parameter $\beta_i$ (that is minimized at every epoch) and the resulting $L_0$ sparsity (see Equation \ref{eq:l0_norm}). Hence, a simple budget-aware extension of our current DAM formulation would be to start from a large value of lambda (to provide sufficient scope for aggressive pruning) and keep monitoring the level of sparsity at every layer during the training process. As soon as the target sparsity mark is achieved, we can freeze $\beta_i$'s at every layer thus essentially halting the pruning process (note that the network architecture becomes immutable if $\beta_i$'s is fixed).

\textbf{DAM Variants for Pre-trained Networks:} One of the fundamental assumptions of our current DAM formulation is that all neurons are symmetrically invariant to each at initialization, and hence we can simply use random neuron ordering for discriminative masking. While this assumption is valid for ``training from scratch'', this may not hold for neurons in a pre-trained network. Future extensions of DAM can include advanced ways of ordering neurons that do not rely on the above assumption and hence can even be used with pre-trained networks. For example, we can order neurons based on the magnitudes of weights of every neuron or the value of mutual information between neuron activations and outputs. This would open novel avenues of research in structured network pruning by finding effective neuron orderings for different initializations of network weights.

\bibliographystyle{plainnat}
\bibliography{main}

\FloatBarrier

\newpage
\maketitle
\appendix

\section{Analysis of the DAM Gate Function Dynamics During Training}
In this section, we theoretically analyze the dynamics
of the DAM mask $\bm{g_i}$ at the $i$-th layer as the training process unfolds. This can be characterized by the movement of the gate function (or gate) during the training process, which is solely dominated by changes in the learnable offset parameter $\beta_i$.

\subsection{Gradients of DAM Learning Objective w.r.t. $\beta_i$}
Building on the notations introduced in Section 3 of the main paper, let us denote the mask value $g_{ij}$ at the $j$-th neuron at $i$-th layer as a function of $\beta_i$, i.e., $\xi_j: \beta_i \to g_{ij}$, where 
\begin{align}
    g_{ij} = \xi_j(\beta_i) & = \max\left[\tanh\left(\alpha_i \left(kj/n_i + \beta_i\right)\right),0\right]
    \label{eq:xi}
\end{align}
Let $f$ denote the continuous function expressed by our neural network with learnable parameters $\Theta$. The loss function for training the neural network for the target task can then be denoted as $L = {\mathcal{L}(f(x, \Theta, \beta_i))} $ (e.g., cross-entropy loss for supervised structured pruning problems and reconstruction error for representation learning problems), where $x$ denotes the input features to the neural network.
Using gradient descent methods with a learning rate of $\eta$, the expected update formula of $\beta_i$ in DAM is given by:
\begin{align}
    \Delta \beta_i & = -  \eta \, \mathbb{E}_{x \sim \mathcal{D}_{tr} } \left[ \nabla_{\beta_i} {\mathcal{L}(f(x, \Theta, \beta_i))} + \lambda \nabla_{\beta_i} {\beta_i/({l-1})}  \right]\\
     & = -  \eta \, \mathbb{E}_{x \sim \mathcal{D}_{tr} } \left[ \nabla_{\beta_i} {\mathcal{L}(f(x, \Theta, \beta_i))} \right] - \eta \lambda/({l-1})
    \label{eq:delta_beta} 
\end{align}
Let $\bm{h_i}$ be the layer output before applying the DAM mask, and the masked output be represented as $\bm{o_i} = \bm{h_i} \circ \bm{g_i}$ after applying the gate. The gradient of the loss w.r.t. $\beta_i$ can be obtained by applying the chain rule of differentiation as follows:
\begin{align}
    \nabla_{\beta_i} {\mathcal{L}(f(x, \Theta, \beta_i))} 
    = \frac{\partial {\mathcal{L}(f(x, \Theta, \beta_i))}}{\partial \bm{o_i}} 
    \frac{\partial \bm{o_i}}{\partial \bm{g_i}}
    \frac{\partial \bm{g_i}}{\partial \beta_i}
    = \sum_{j=1}^{n_i} \frac{\partial {\mathcal{L}(f(x, \Theta, \beta_i))}}{\partial {o_{ij}}} 
    \frac{\partial {o_{ij}}}{\partial {g_{ij}}}
    \frac{\partial {g_{ij}}}{\partial \beta_i}
    \label{eq:chain_rule}
\end{align}
Let us analyze this gradient by observing the last term of this equation, $\partial {g_{ij}} / \partial \beta_i$. For the $j$-th neuron, $\partial g_{ij} /\partial \beta_i = 0$ if and only if $\partial \xi_j(\beta_i) /\partial \beta_i = 0$. Since $tanh(z)$ has non-zero gradients for $z > 0$, the gradient of $\xi_j(\beta_i)$ is 0 only when $kj/n_i + \beta_i \leq 0$, i.e., the 
mask value of the neuron is 0 (or in other words, it is \textit{deactivated} or dead). Let us denote the set of all neuron indices with non-zero mask values (also referred to as \textit{active} neurons) as $\mathcal{J}$. Equation \ref{eq:chain_rule} can then be simplified as:
\begin{align}
    \nabla_{\beta_i} {\mathcal{L}(f(x, \Theta,\beta_i))} 
    & = \alpha_i \sum_{j \in \mathcal{J}} \underbrace{\frac{\partial {\mathcal{L}(f(x, \Theta, \beta_i))}}{\partial o_{ij}} h_{ij} \left( 1 - g_{ij}^2 \right)}_{q_{ij}},
    \label{eq:derivative_beta_obj} \\
    & = \alpha_i \sum_{j \in \mathcal{J}} {q_{ij}}, \label{eq:derivative_beta_obj_simplified}
\end{align}
where $q_{ij}$ represents the contribution of the $j$-th neuron to the gradient of the loss term with respect to $beta_i$. We can make the following two observations: (i) only those neurons that are active (i.e., have non-zero mask values) have a contribution towards updating $\beta_i$ and moving the gate function. (ii) If the mask value of a neuron is 1 (i.e., $g_{ij} = 1$), then their contribution to the gradient of the loss w.r.t. $\beta_i$ is again 0.
It shows that the neurons that play an important role in the dynamics of the gates are the ones with non-zero activation $h_{ij}$ and mask values that have not saturated to 1 (i.e., $g_{ij} < 1$). We name these neurons as \textit{support} neurons and their position in the ordering of neurons as the \textit{transitioning zone} of the gate function. Similarly, neurons with zero mask values are termed as \textit{deactivated} neurons and the neurons with mask values close to 1 as \textit{privileged} neurons (since they are never turned off).

\subsection{Equilibrium of DAM Gate Function upon Convergence}
We next study the properties of the equilibrium solution of $\beta_i$ that we arrive upon convergence of the DAM training process. Suppose that we have converged to  $\beta_i^*$ for the offset parameter for the $i$-th layer. This would mean that the gradient of the DAM learning objective w.r.t. $\beta_i^*$ would be equal to 0 as follows:
\begin{align}
    \nabla_{\beta_i^*} {\mathcal{L}(f(x, \Theta, \beta_i^*))} + \lambda/(l-1) = 0 
\end{align}
Substituting the value of $\nabla_{\beta_i} {\mathcal{L}(f(x, \Theta, \beta_i))}$ from Equation (\ref{eq:derivative_beta_obj_simplified}) and rearranging terms, we get
\begin{align}
    \sum_{j \in \mathcal{J}} q_{ij} = - \lambda / (\alpha_i(l-1)). \label{eq:equilibrium}
\end{align}

Since $\lambda, \alpha_i > 0$, the equilibrium exists only when the sum of $q_{ij}$ is negative. This happens when decreasing the masked outputs $o_{ij}$ of the support neurons in the transitioning zone leads to an increase in the loss function, signifying that any further pruning can lead to loss of accuracy. In other words, the training dynamics of the gate function stops when the features learned at the support neurons are useful enough that their pruning is detrimental to the generalization performance of the network.

\subsection{Additional Remarks on the Effects of $\alpha_i$ and $\lambda$ On DAM Convergence}
Equation \ref{eq:equilibrium} also implies that a large value of $\alpha_i$ or a low value of $\lambda$ may make the equilibrium easy to reach, thus leading to \textit{premature convergence} to a pruned network that has not been fully trained to capture refined features at its support neurons. This is supported by our empirical observations that large values of $\alpha_i$ tends to prevent DAM layers to reach higher sparsity. We thus set $\alpha_i = 1$ in all our experiments. On the other hand, the choice of $\lambda$ also plays a big role as we observed in our experiments (e.g., as is shown in Section C via hyperparameter sensitivity analysis).

Further note that the masking dynamics of DAM involve a smooth transition from an unpruned network to a pruned network where at every training epoch, some of the neurons (or channels) gradually die out while the rest remain unaffected. The neurons at the edge of being dropped out are gradually assigned low gate values such that the other neurons can slowly re-adapt themselves to pick up or recover the features that are being dropped out.
However, steep gate function choices (that have narrow transitioning zones) may cause this transition to be too fast such that the network suffers a drop in accuracy as neurons on the edge are dropped while the other neurons do not have sufficient time to recover the dropped features.
In our formulation, choosing a very large value of $\alpha_i$ can make the gate function too steep for effective pruning (we thus choose $\alpha_i = 1$). For other alternate choices of gate function, $\xi_j(\beta_i)$, than what we used in our current implementation, similar considerations need to be observed to avoid the gate function from becoming too steep. 

\subsection{Alternative Choices of Gate Function}
In our experiments, we also tried using the sigmoid as the gate function in DAM. 
We found that sigmoid cannot be used as a valid choice for the gate function as it does not have the ``hard-thresholding'' property, i.e., the gate function value of sigmoid does not remain exactly 0 before a certain threshold value of input is reached. This property is important to enforce strict $L_0$ sparsity, otherwise the neurons would still remain active with small non-zero gate values across a large range of inputs. In general, there are three properties that we desire in an ideal gate function: (a) it should be exactly 0 before reaching a certain threshold value of input, (b) it should be monotonically increasing, and (c) it should have a parameter to control the steepness of the gate function. While we found ReLU-tanh to exhibit stable training dynamics across different datasets in our experiments, other choices of gate functions can also be explored (e.g., hard-sigmoid).

\section{Theoretical Results And Technical Proofs For Section 4}

We empirically observed in Section 4 of the main paper that for the linear dimensionality reduction case, we always converged to the same solution for a constant setting of hyper-parameters regardless of the random initialization of the neural network. To theoretically understand if DAM is capable of converging to the optimal solution (where the pruned bottleneck dimension is exactly same as the rank of the input data matrix), we provide further theoretical analysis for the ability of DAM to perform linear dimensionality reduction (DR) in the following.

\textbf{Problem Statment}:
Let $G = \diag(\bm{g})$, where $\bm{g} \in \mathbb{R}^n$ is the vector of mask values at the bottleneck layer given by
\begin{equation}
    \bm{g} = \max\left[ \tanh(\bm{\mu} + \beta \bm{1}), 0 \right]
\end{equation}
and the $j$-th element in $\bm{\mu}$ is given by $\mu_j = kj/n$, while $\beta \in \mathbb{R}$ is a scalar learnable parameter.
Let the real matrices $A_1, A_2$ define two linear transformations $A_1: \mathbb{R}^n \longrightarrow \mathbb{R}^d$ and $A_2: \mathbb{R}^d \longrightarrow \mathbb{R}^n$. Let us consider a bounded input data matrix, $X \in \mathbb{R}^{d \times N}$, that is rank-deficient (with rank $m$), where $N$ is the number of data samples and $d$ is the number of features, $N \gg d > m$. Then, the input $X$ and its reconstruction $\hat{X}$ are related by the following equation:
\begin{equation}
    \hat{X} = A_1 G A_2 X.
    \label{eq:linear_dam}
\end{equation}
Let $\Theta = \{A_1, A_2, \beta\}$ be the set of learnable parameters. For a given $X$, let us define the multi-objective minimization problem as 
\begin{align}
    \min_{\Theta}\left( \norm{\hat{X} - X}_F^2 , \beta \right),
    \label{eq:linear_dam_objective}
\end{align}
where $\norm{.}_F^2$ denotes the Frobenius norm of a matrix. We are interested in reaching a Pareto optimal solution with a predefined trade-off hyper-parameter between the two objectives using the gradient descent algorithm. 

\textbf{Optimization Scheme}:
Let $L = \norm{\hat{X} - X}_F^2 + \lambda \beta$ and initial values for the learnable parameters be $A_1^{(0)}, A_2^{(0)}, \beta^{(0)}$, where $\lambda > 0$ is the trade-off hyper-parameter. 
Using gradient descent (GD), for every parameter $\theta \in \Theta$, the updating rule for the $t$-th iteration can be written as
\begin{align}
    \theta^{(t+1)} = \theta^{(t)} - \eta(\grad_{\theta} L)^{(t)}
    \label{algo: gd_linear}
\end{align}
where $\eta \in \mathbb{R}^+$ is the step size. We assume that we allow the network to perform sufficient number of iterations of gradient updates before it reaches convergence.

\begin{theorem}
(Existence of Optimal Solution) Assuming all the $m$ singular values of $X$ are non-trivial (i.e., they are all greater than some positive value $\epsilon$), there exists $\lambda > 0$ such that at the minima solution of $\mathcal{L}$, $\hat{X} = X$ and the number of non-zero entries of $\bm{g}$ equals $m$.
\label{thm: accuracy}
\end{theorem}
\textit{Proof:}\ \ Let $\Tilde{A_2} = G A_2$. Then, we can rewrite $\hat{X}$ as a function of $\Tilde{A_2}$ as follows 
\begin{equation}
    \hat{X} = A_1 \Tilde{A_2} X.
\end{equation}
Using this value of $\hat{X}$ as a function of $\Tilde{A_2}$, we can decouple the combined learning objective into the following two optimization problems:
\begin{align}
    \left( \min_{\{ A_1, \Tilde{A_2} \}}\norm{A_1 \Tilde{A_2} X - X}^2_F, ~~ \min_{\beta} {\lambda \beta}\right),
    \label{eq:reparam_objective}
\end{align}
where the first term only depends on the variables $A_1, \Tilde{A_2}$, while the second term only depends on $\beta$. Optimizing the second term is trivial since it is a linear function of $\beta$. On the other hand, the first term has its minimum value at 
\begin{align}
    A_1 \Tilde{A_2} X &= X \notag \\ 
    A_1 &= \Tilde{A_2}^+ \label{eq:optimal_A}
\end{align}
where $\Tilde{A_2}^+$ is the \textit{Moorse-Penrose Inverse} of $\Tilde{A_2}$. 
Moreover, whenever $A_1 = \Tilde{A_2}^+$, the following relationship would also hold
\begin{equation}
    \rank(\Tilde{A_2}) \geq \rank(A_1 \Tilde{A_2} X) = \rank(X) = m.
\end{equation}
and $\rank(\Tilde{A_2}) = \rank(G A_2) \leq \rank(G)$. Thus, $\rank(G) \geq m$. Assuming the diagonal matrix $G$ has $m'$ nonzero entries, then $m' = \rank(G) \geq m$. This means that the lowest value that $m'$ can take is $m$, and we want to study if there exists some setting of $A_1, \Tilde{A_2}$, and $\beta$ that leads to $m' = m$. We specifically explore the setting where the rank of $G$ (the number of non-zero values of $\bm{g}$) is equal to $m$. Note that in our DAM formulation, the number of non-zero entries of the mask vector $\bm{g}$ is directly related to $\beta$ as follows (see Section 3 of main paper for more details):
\begin{equation}
    m' = \norm{G}_0 = \ceil{n(1 + \beta /k)} = m,
\end{equation}
\begin{figure}[t]
    \centering
    \includegraphics[width=0.35\textwidth]{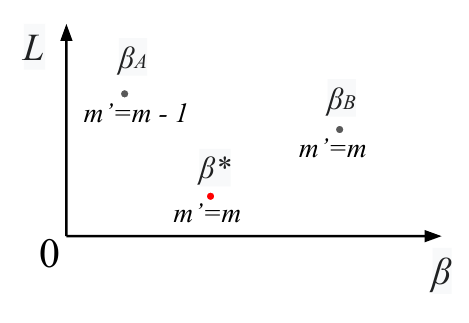}
    \caption{Schematic plot showing variations in the learning objective value $L$ as a function of $\beta$ near the minima of $L$ (shown as red $\beta^*$).}
    \label{fig:thm1}
\end{figure}
which means that if we remove the ceiling operator, the following inequality holds:
\begin{align}
    m &\geq n(1 + \beta /k)> m - 1\\
    k \left( \frac{m}{n} - 1 \right) &\geq \beta > k \left( \frac{m - 1}{n} - 1 \right)
\end{align}
As illustrated in Figure 1, we seek to show that a critical point of $L$ in terms of $\beta$ exists such that $m'= m$. Let $\beta_A = k \left( \frac{m - 1}{n} - 1 \right), \beta_B = k \left( \frac{m}{n} - 1 \right)$ and denote their corresponding values of $L$ as $L_A$ and $L_B$ respectively. As we showed in previous discussions, with $m' \geq m$, as long as $A_1 = \Tilde{A_2}^+$, then the first term of $L$ is 0. Thus, there exists a $\beta' < \beta_B$ such that $m' = m$ ($\beta' > \beta_A$), and its corresponding value of $L$ is smaller than $L_B$ (because of a smaller second term). On the other hand, we want to discuss whether the optimization over Eq. (\ref{eq:reparam_objective}) will lead to a smaller $\beta$ such that $G$ has more nonzero entries (consequently making the second term of Eq. (\ref{eq:reparam_objective}) smaller). In particular, when $m' < m$, then 
\begin{equation}
    \rank(\hat{X}) = \rank(A_1\Tilde{A_2}X) \leq \rank(\Tilde{A_2}) \leq \rank(G) \leq m' < m.
\end{equation}
Since $X$ has no trivial singular values, when $\rank(\hat{X}) < m$, then $\norm{\hat{X} - X}_F^2$ would also take non-trivial values, i.e., $\norm{\hat{X} - X}_F^2 > c\epsilon$, where $c$ is some positive constant. Thus, assuming $\lambda$ is reasonably small, there exists a $\beta' > \beta_A$ such that $m' = m$ ($\beta' \leq \beta_B$), and its corresponding value of $L$ is smaller than $\beta_B$'s corresponding value of $L_A$ (because of a smaller first term).

Thus, there exists a critical point of $L$ (denoted as $\beta^*$) between $\beta_A$ and $\beta_B$, which is the minima of $L$. For $\beta^*$, we have $m' = m$. Therefore we have proven the existence of an optimal solution of DAM. 
In summary, the optimal set is
\begin{align}
\begin{split}
    \Theta^* = \bigg \{A_1, A_2, \beta | &\forall A_1 \in \mathbb{R}^{d \times n}, \forall A_2 \in \mathbb{R}^{n \times d}, \forall \beta \in \mathbb{R}, \textup{s.t.}\\
      &A_1 = (G A_2)^+ ,k \left( \frac{m}{n} - 1 \right) \geq \beta > k \left( \frac{m - 1}{n} - 1 \right) \Bigg\}
\end{split}
\end{align}

\begin{figure}[ht]
\begin{subfigure}{0.9\textwidth}
  \centering
  \includegraphics[width=\linewidth]{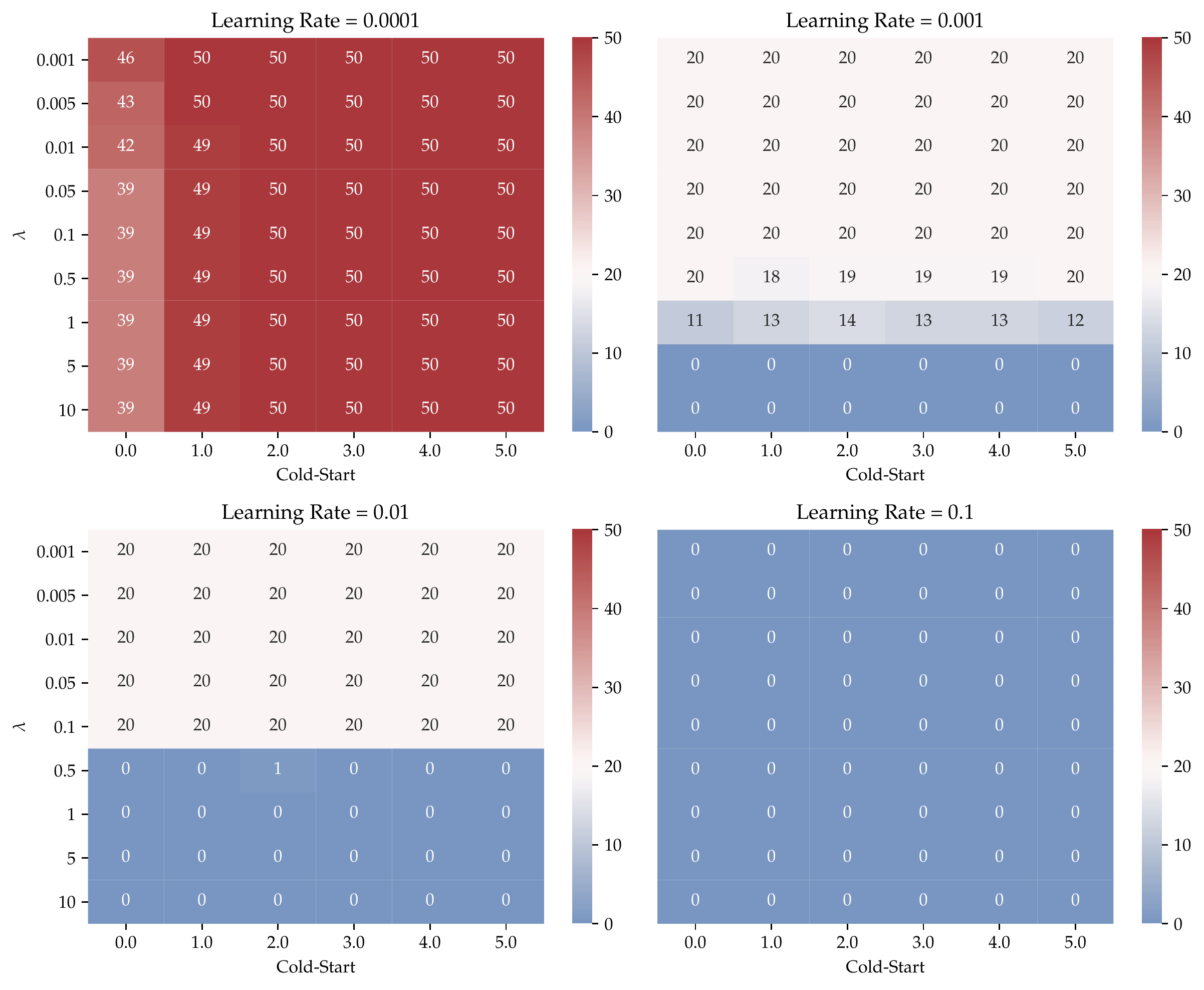}
  \caption{Bottleneck Dimension.}
  \label{fig:btl-lin}
\end{subfigure}
\begin{subfigure}{0.9\textwidth}
  \centering
  \includegraphics[width=\linewidth]{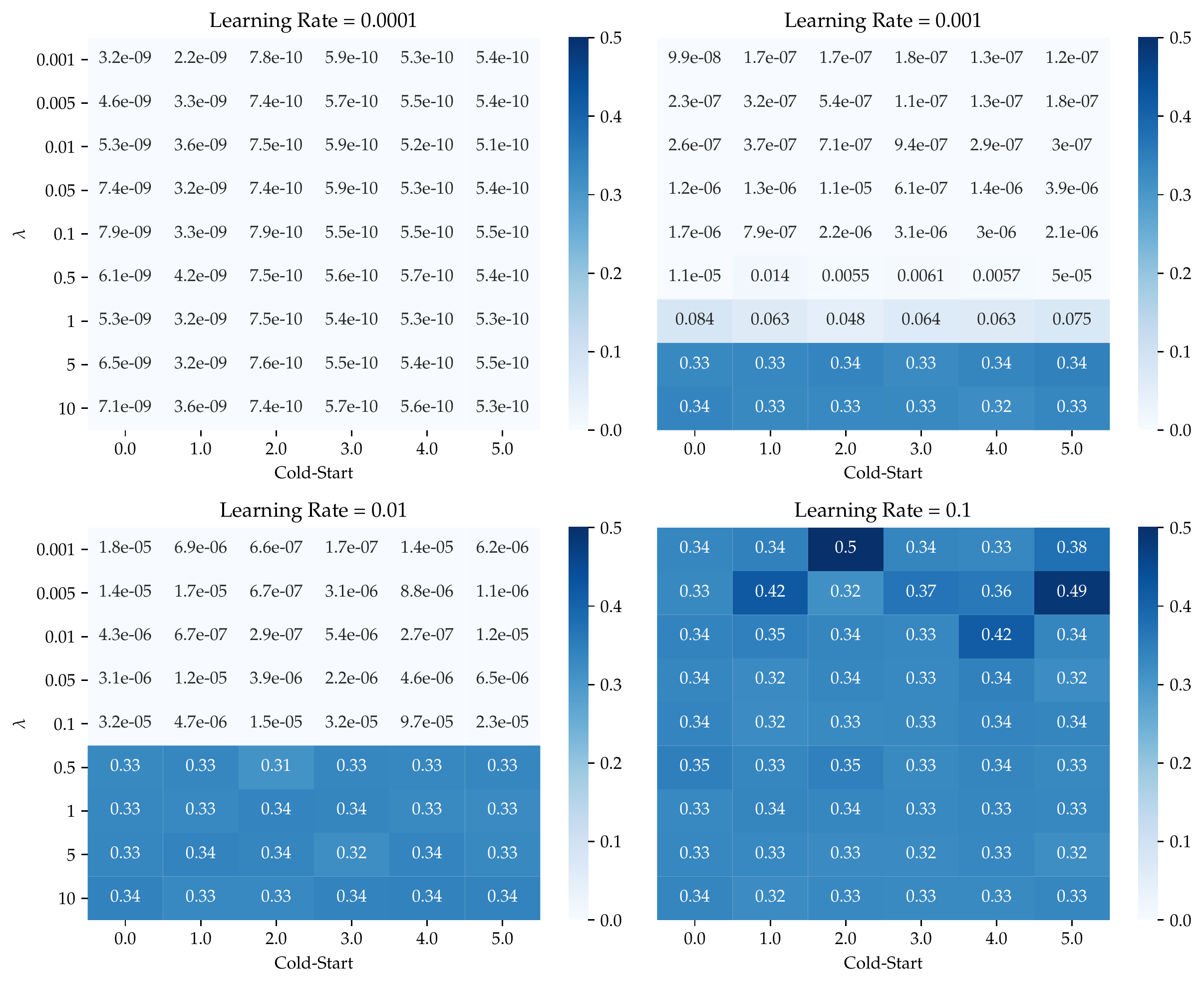}
  \caption{Reconstruction Loss.}
  \label{fig:rec-lin}
\end{subfigure}
\caption{
Hyper-parameter sensitivity of DAM for dimensionality reduction (Section 4, Linear).
}
\label{fig:sensitivity-linear}
\end{figure}

\begin{figure}[ht]
\begin{subfigure}{0.9\textwidth}
  \centering
  \includegraphics[width=\linewidth]{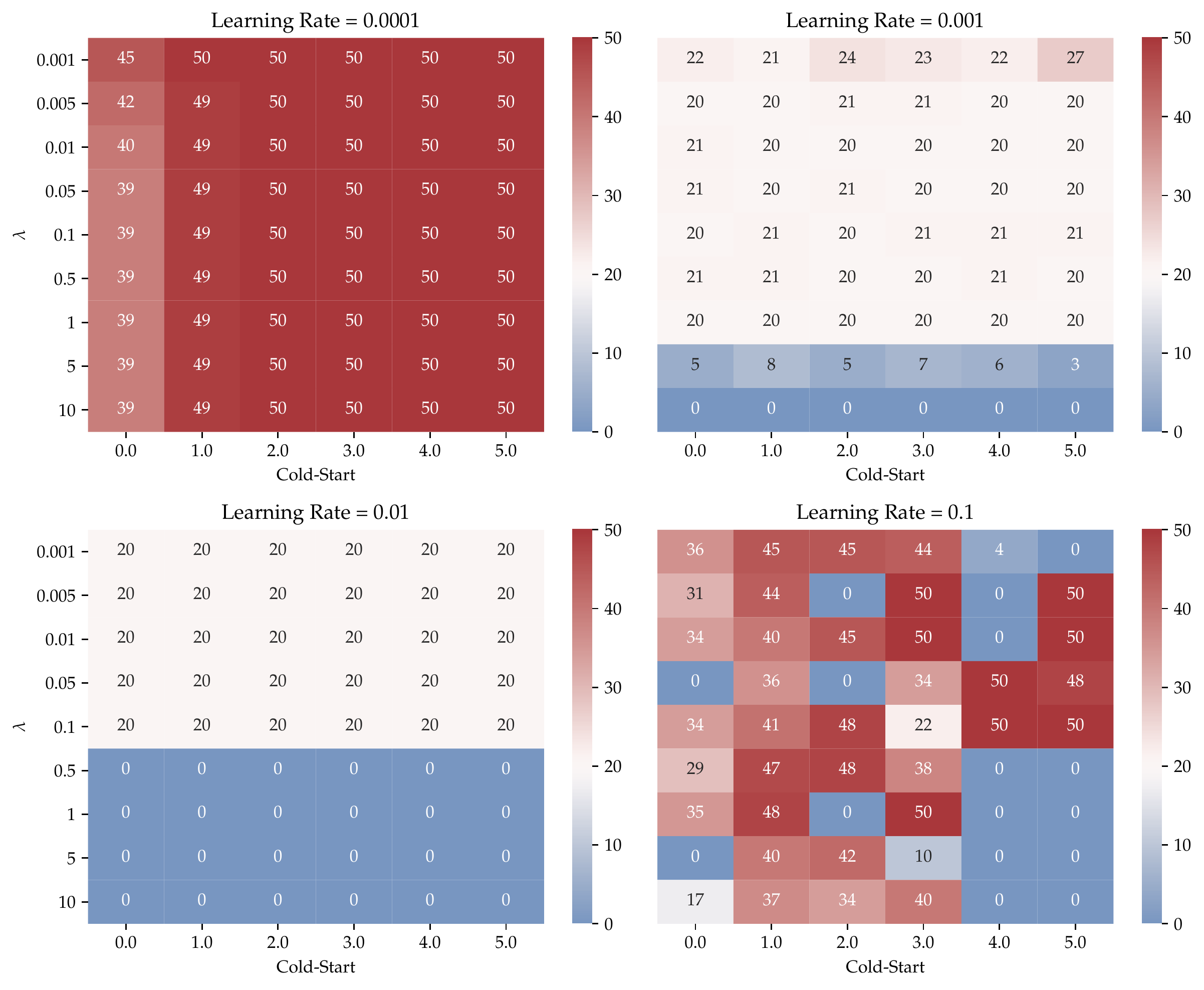}
  \caption{Bottleneck Dimension.}
  \label{fig:btl-quad}
\end{subfigure}
\begin{subfigure}{0.9\textwidth}
  \centering
  \includegraphics[width=\linewidth]{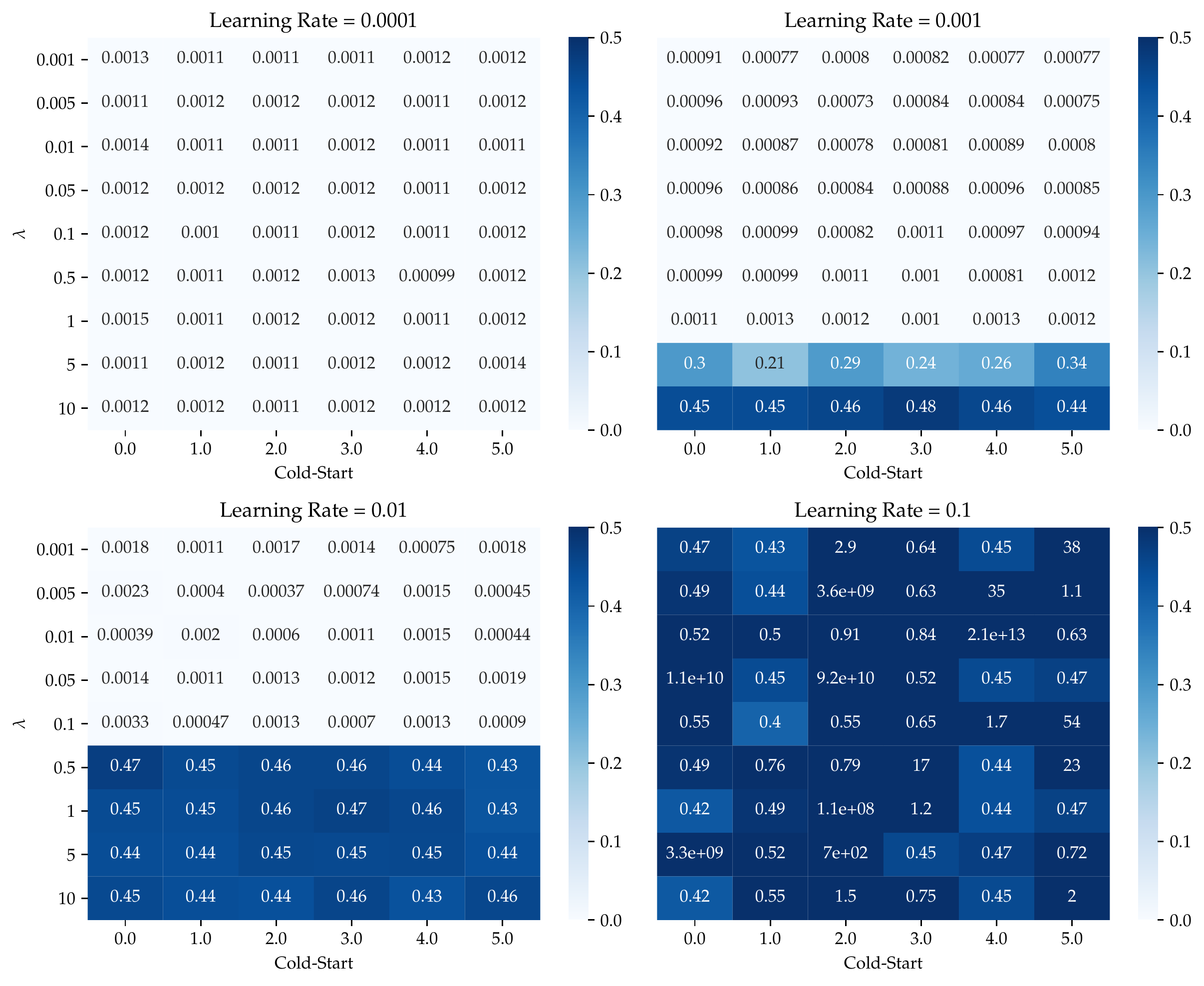}
  \caption{Reconstruction Loss.}
  \label{fig:rec-quad}
\end{subfigure}
\caption{
Hyper-parameter sensitivity of DAM for dimensionality reduction (Section 4, Polynomial).
}
\label{fig:sensitivity-quad}
\end{figure}

\begin{figure}[ht]
\begin{subfigure}{0.9\textwidth}
  \centering
  \includegraphics[width=\linewidth]{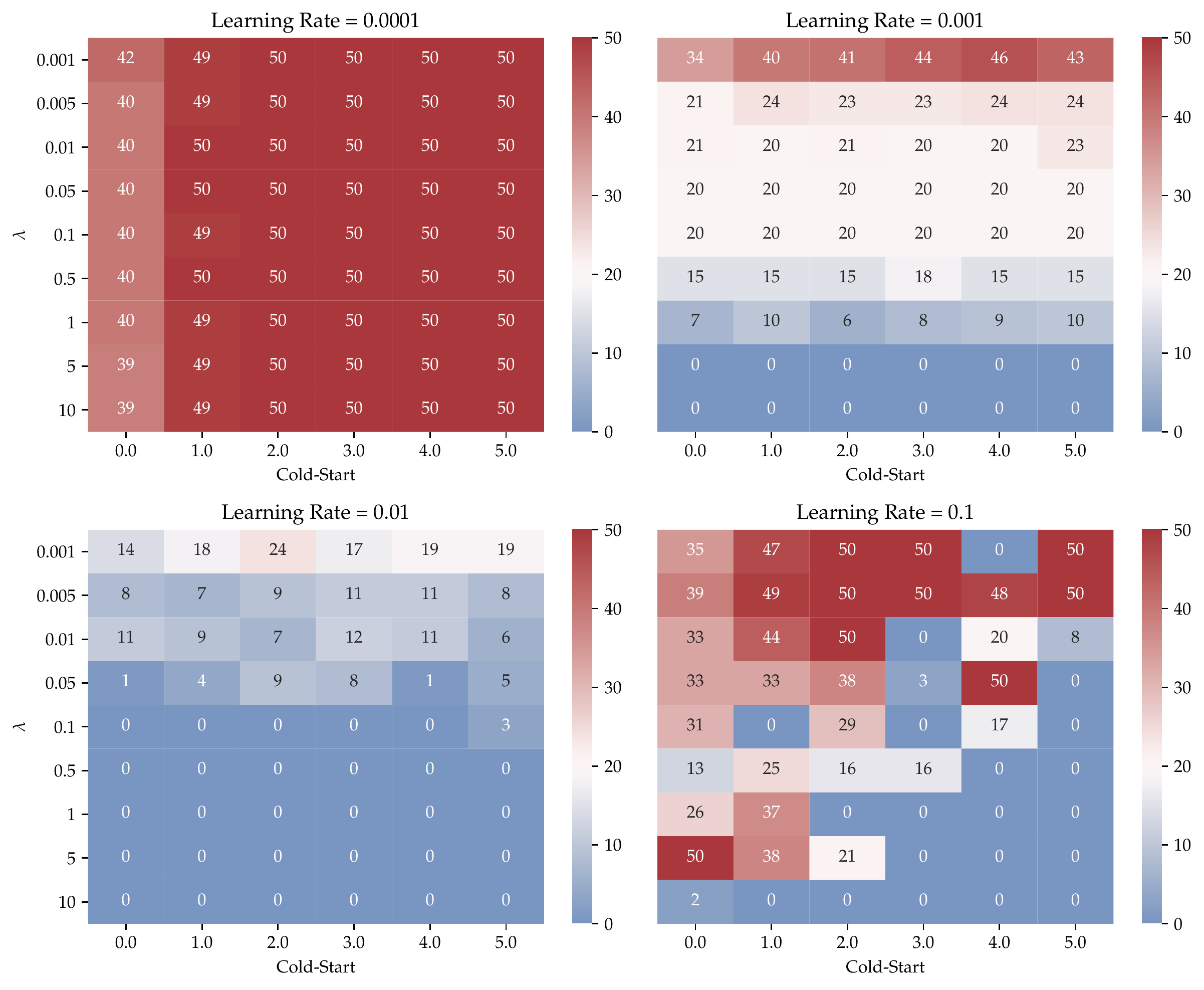}
  \caption{Bottleneck Dimension.}
  \label{fig:btl-nn}
\end{subfigure}
\begin{subfigure}{0.9\textwidth}
  \centering
  \includegraphics[width=\linewidth]{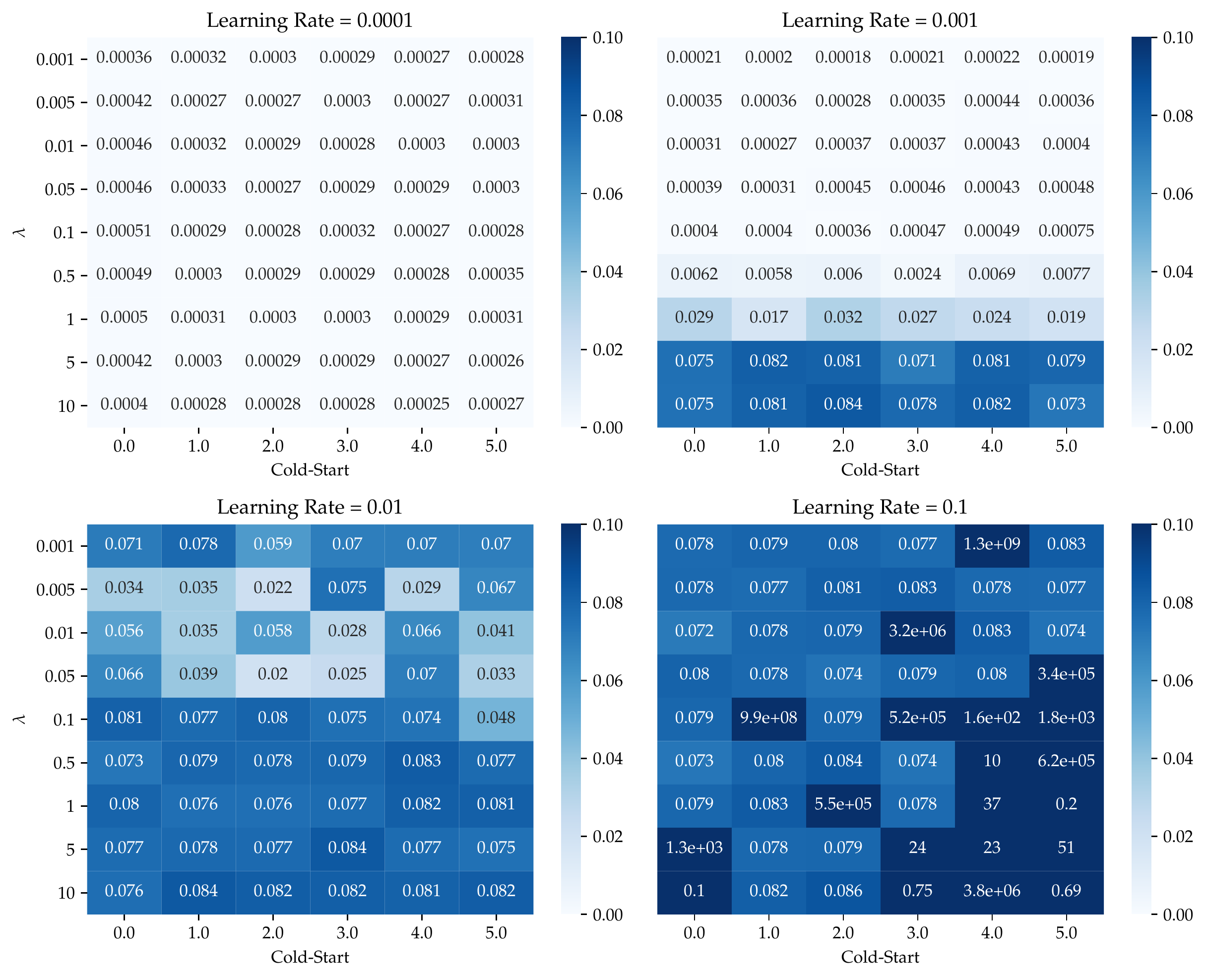}
  \caption{Reconstruction Loss.}
  \label{fig:rec-nn}
\end{subfigure}
\caption{
Hyper-parameter sensitivity of DAM for dimensionality reduction (Section 4, Neural Network).
}
\label{fig:sensitivity-nn}
\end{figure}
\section{Hyperparameter Sensitivity}
We have demonstrated that DAM performs very well in the representation learning experiments for dimensionality reduction. Here, we study how stable DAM is with respect to variations in hyperparameter settings.
We specifically examine three hyperparameters: learning rate, trade-off parameter, $\lambda$, and the initial value of $\beta$, $\beta_0$. The results of this experiemnt are shown in the form of heatmaps of bottleneck dimension and reconstruction loss upon convergence in Figures \ref{fig:sensitivity-linear}, \ref{fig:sensitivity-quad} and \ref{fig:sensitivity-nn}. 

We can observe that as we vary the learning rate from 0.0001 to 0.1 (which represents a range covering three orders of magnitude), the results for all three cases show variations, where the best results are obtained in the medium range of values (from 0.001 to 0.01), which are common choices of learning rates used in conventional deep learning frameworks. This variation of our results with learning rate is common to any other algorithm based on gradient descent methods. However, for a fixed learning rate,  we can see that the results of DAM are quite stable to the choice of $\lambda$ and $\beta_0$. For example, we can see that the results of DAM are consistent across different choices of cold-start ($\beta^{(0)}$) for all choices of learning rates in the three cases. The sensitivity in terms of $\lambda$ is also low but increases as the feature correlations gets more and more complicated, i.e., as we transition from linear to quadratic to neural network cases. 
In fact, we observe that small $\lambda$ values result in slower or premature convergence, while too large $\lambda$ can break the model by pruning out all of the parameters. In summary, we can say that by using common choices of learning rate (e.g., 0.01 or 0.001), DAM is able to produce consistent results with reasonable choice of $\lambda$ (e.g., $\lambda = 0.01$).

\section{Experimental Setups}
\subsection{Representation Learning (Section 4)}
\textbf{Datasets:}
For dimensionality reduction experiments, we created three synthetic datasets for each of the mapping functions. To generate each dataset, we used two matrices, one as as an encoder and another one as a decoder. For the linear case, the the two matrices were initialized with values sampled from a normal distribution. For the quadratic case, we used a three-layer MLP as the encoder and one QRes layer \cite{jie2021qres} as the decoder for the quadratic mapping function. For the neural network case, we used a three-layer MLP as the encoder and two-layer MLP as the decoder. 
We varied the underlying factor ($r$) from 5 to 20 to generate these datasets. 
For the recommendation system experiments, we used three benchmark datasets Flixter \cite{jamali2010matrix}, Douban \cite{ma2011recommender}, and Yahoo-Music \cite{dror2012yahoo}.
For graph representation learning, we used three popular citation networks: Cora, CiteSeer, and PubMed \cite{yang2016revisiting}. 
We further used the MNIST \cite{lecun1998mnist} dataset to compare between DAM and $L_{1}$-norm based auto-encoders. 

\begin{table}[t]
\begin{center}
\caption{Experiment Setups And Implementation Details For Representation Learning (Section 4).}
\begin{small}
\vspace{2ex}
\begin{tabular}{lllllcc}
\thickhline
Source  & Dataset        & Model          & Learning Rate             & Epochs  & $\lambda$  & Cold-Start\\ \hline
Fig. 3a & Synthetic      & Linear    & 0.01, \:$L_2$ = $10^{-6}$ & 2,000 & 0.01 & $\beta^{(0)} = 1$, 0 epochs \\
Fig. 3b & Synthetic      & Quadratic & 0.01, \:$L_2$ = $10^{-6}$ & 5,000 & 0.01  & $\beta^{(0)} = 5$, 0 epochs \\
Fig. 3c & Synthetic      & NN        & 0.001, $L_2$ = $0$        & 10,000 & 0.1  & $\beta^{(0)} = 1$, 0 epochs \\ \hline
Tab. 2  & Flixter        & IGMC           & 0.001, $L_2$ = $0$         & 100 & -  & - \\
Tab. 2  & Flixter        & IGMC-DAM       & 0.001, $L_2$ = $0$         & 100 & 0.1  & $\beta^{(0)} = 1$, 0 epochs \\
Tab. 2  & Douban         & IGMC           & 0.001, $L_2$ = $0$         & 100 & -  & - \\
Tab. 2  & Douban         & IGMC-DAM       & 0.001, $L_2$ = $0$         & 100 & 0.1  & $\beta^{(0)} = 1$, 0 epochs \\
Tab. 2  & Yahoo-Music    & IGMC           & 0.001, $L_2$ = $0$         & 100 & -  & - \\
Tab. 2  & Yahoo-Music    & IGMC-DAM       & 0.001, $L_2$ = $0$         & 100 & 50.0  & $\beta^{(0)} = 1$, 0 epochs \\ \hline
Fig. 4a & Cora           & GAE-DAM        & 0.01, \:$L_2$ = $0$       & 1,000 & variable  & $\beta^{(0)} = 1$, 0 epochs \\ 
Fig. 4a & CiteSeer       & GAE-DAM        & 0.01, \:$L_2$ = $0$       & 1,000 & variable  & $\beta^{(0)} = 1$, 0 epochs \\ 
Fig. 4a & PubMed         & GAE-DAM        & 0.01, \:$L_2$ = $0$       & 1,000 & variable  & $\beta^{(0)} = 1$, 0 epochs \\ \hline
Fig. 5 &  MNIST          & $L_{1}$-norm   & 0.001, \:$L_2$ = $0$      & 100 & variable  & - \\ 
Fig. 5 &  MNIST          & DAM            & 0.001, \:$L_2$ = $0$          & 100 & variable  & $\beta^{(0)} = 1$, 0 epochs \\ \hline
\thickhline      
\end{tabular}
\label{tab:specifications_CRL}
\end{small}
\end{center}
\end{table}

\textbf{DAM Implementation Specifications:}
\begin{itemize}
\item Table 1 provides full details of the hyper-parameter choices and implementation details of DAM used for generating all of the results in representation learning experiments of Section 4. We used Adam optimizer for all these experiments.
\item All our results are reported by taking the mean and standard deviation over five independent random runs. 
\item We used the linear activation function for the linear dimensionality reduction experiments, LeakyRelu activation function for the quadratic experiments, and the ELU activation function for the neural network mapping experiments. Note that non-linear activation functions were required for the quadratic and neural network experiments, because of the non-linearity of the problem. We added a DAM layer between the encoder and decoder models to perform dimensionality reduction on the bottleneck layer.
\item We use the official implementation of IGMC \footnote{\url{https://github.com/muhanzhang/IGMC}} to run the recommendation systems experiments on Flixter, Douban, and Yahoo-Music datasets. 
\item We use the PyTorch Geometric implementation of GAE \footnote{\url{https://github.com/rusty1s/pytorch_geometric/blob/master/examples/autoencoder.py}} for our graph representation learning experiments.
\item We kept the same architecture as the official implementations of IGMC and GAE for all the recommendation system and graph representation learning experiments. For IGMC-DAM and GAE-DAM, we simply added one DAM layer after the bottleneck layer to perform pruning of the bottleneck dimension. 
\item For the dimensionality reduction experiments on the MNIST dataset, we used a plain autoencoder setup. We used four layers with 784, 64, 32, 50 neurons for the encoder, and for the decoder, we used four layers with 50, 32, 64, and 784 neurons. ReLU was used as the activation function. After the encoder layer, we used a DAM layer for dimensionality reduction. Likewise, we append an $L_1$-norm layer after the encoder layer for the $L_1$-norm-based dimensionality reduction.
\item We varied $\lambda$ values for the graph representation learning experiments and MNIST dataset as follows. We varied the $\lambda$ from 0.01 to 5 for both GAE and GAE-DAM for graph representation learning experiments. We varied the $\lambda$ from 0.01 to 10 for the DAM model on the MNIST dataset.  Likewise, we changed the range of $\lambda$ from 0.01 to 25 for the $L_1$-norm model. 
\item We observed some improvement in the performance of the baseline IGMC model for 100 epochs, although the official implementation used 40 epochs. We kept all other hyperparameters same as the official implementation. 
\end{itemize}

\subsection{Structured Network Pruning (Section 5)}
\begin{table}[t]
\begin{center}
\caption{Experiment Setups For Structured Network Pruning (Section 5).}
\begin{small}
\vspace{2ex}
\begin{tabular}{lllllcc}
\thickhline
Source  & Dataset        & Model          & Learning Rate             & Epochs  & $\lambda$  & Cold-Start\\             \hline
Fig. 6  & CIFAR-10       & DAM            & 0.05, \:$L_2$ = $10^{-3}$ & 200+0+0 & variable & $\beta^{(0)} = 1$, 20 epochs   \\
Fig. 6  & CIFAR-10       & Net-Slim       & 0.05, \:$L_2$ = $10^{-3}$ & 200+0+50 & -  & - \\
Fig. 6  & CIFAR-10       & ChipNet        & 0.05, \:$L_2$ = $10^{-3}$ & 200+20+50 & -  & - \\
Fig. 6  & CIFAR-100       & DAM            & 0.05, \:$L_2$ = $10^{-3}$ & 200+0+0  & variable  & $\beta^{(0)} = 1$, 20 epochs \\
Fig. 6  & CIFAR-100       & Net-Slim       & 0.05, \:$L_2$ = $10^{-3}$ & 200+0+50 & -  & - \\
Fig. 6  & CIFAR-100       & ChipNet        & 0.05, \:$L_2$ = $10^{-3}$ & 200+20+50 & -  & - \\\hline
Fig. 7  & CIFAR-10       & DAM            & 0.05, \:$L_2$ = $10^{-3}$ & 200+0+0  & 0.4  & $\beta^{(0)} = 1$, 20 epochs \\
Fig. 7  & CIFAR-10       & Net-Slim       & 0.05, \:$L_2$ = $10^{-3}$ & 200+0+50 & -  & - \\
Fig. 7  & CIFAR-10       & ChipNet        & 0.05, \:$L_2$ = $10^{-3}$ & 200+20+50 & -  & - \\\hline
Tab. 2  & CIFAR-10       & DAM            & 0.05, \:$L_2$ = $10^{-3}$ & 200+0+0  & 0.4  & $\beta^{(0)} = 1$, 20 epochs \\
Tab. 2  & CIFAR-10       & Net-Slim       & 0.05, \:$L_2$ = $10^{-3}$ & 200+0+50 & -  & - \\
Tab. 2  & CIFAR-10       & ChipNet        & 0.05, \:$L_2$ = $10^{-3}$ & 200+20+50 & -  & - \\
\thickhline      
\end{tabular}
\label{tab:specifications_NP}
\end{small}
\end{center}
\end{table}
\textbf{DAM Implementation Specifications:}
\begin{itemize}
    \item All the models were trained using SGD optimizer.
    \item Table \ref{tab:specifications_NP} provides full details of the hyperparameter choices and implementation details of DAM for all structured network pruning experiments.
    \item For running time comparisons, we used an \textit{NVIDIA TITAN RTX} graphic card with \textit{Intel Xeon Gold 6240} CPU on \textit{Ubuntu 18.04 LTS} system. The channel pruning ratio for Net-Slim and ChipNet is set to 0.4 in all our experiments.
    \item For pruning experiments, we used learning rate decay as adopted in the ChipNet official implementation\footnote{ \url{https://github.com/transmuteAI/ChipNet}}. Table \ref{tab:specifications_NP} only shows the initial learning rate in these experiments.
    \item For all methods, the epochs are shown in the format of training epochs + pruning epochs + finetuning epochs. We can see that DAM requires 0 pruning epochs and 0 finetuning epochs, Net-Slim has 0 finetuning epochs, whereas ChipNet has non-zero epochs for all three stages.
    \item For Figure 6, on CIFAR-10 we use the following range of $\lambda$ values for DAM, $\lambda=\{0.0, 0.1, 0.2, 0.3, 0.4, 0.5, 0.75\}$. For ChipNet and Net-Slim, we used the following range of values for pruning ratios: $\{0.1, 0.2, 0.4, 0.6, 1.0\}$ and $\{0.1, 0.2, 0.3, 0.4, 0.5, 0.6, 1.0\}$, respectively. For CIFAR-100, we use $\lambda=\{0.1, 0.2, 0.3, 0.4, 0.5, 0.75, 1.0\}$ for DAM, and pruning ratios of $\{0.2, 0.4, 0.6, 1.0\}$ and $\{0.2, 0.3, 0.4, 0.5, 0.6, 1.0\}$ for ChipNet and Network Slimming, respectively. We did not report the $0.1$ pruning ratio on CIFAR-100 since both ChipNet and Network Slimming demonstrated unstable results.
\end{itemize}

\section{Additional Experimental Results}
\subsection{Effect of Gradient Noise and Activation Functions on MNIST Dataset}

\begin{figure}[H]
    \centering
    \includegraphics[width=0.95\textwidth]{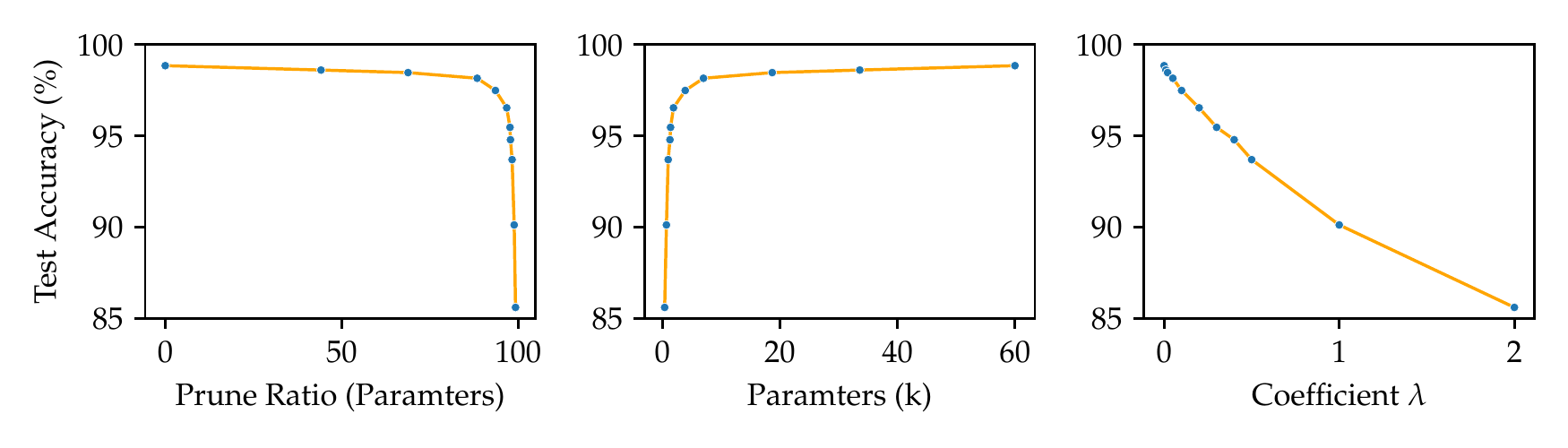}
    \caption{Results of DAM trained with varying $\lambda$ on MNIST dataset. Accuracy is reported using the mean values of 5 random runs.}
    \label{fig:mnist_dam}
\end{figure}

We performed further experiments to evaluate the network pruning performance of DAM using LeNet-5 on MNIST dataset, which is a common dataset for experiments adopted by many previous works. Table \ref{tab:mnist_dam_lenet} (a) shows the results of DAM for pruning the LeNet model at varying values of $\lambda$. Figure \ref{fig:mnist_dam} also shows how test accuracy and pruning ratio varies as we change $\lambda$. We can see that the pruning keeps on continuing even with aggressively large values of $\lambda$ (close to 2), which again demonstrates that DAM does not suffer from the saturation effects of $L_1$-based regularization, as was described in Section 4 of the main paper. In addition to pruning the LeNet, we also examine how adding gradient noise \cite{daneshmand2018escaping} at every epoch of neural network training affects the results. Table \ref{tab:mnist_dam_lenet} (b) shows that by adding gradient noises to the training process leads to slight improvements in accuracy and pruning fractions, implying that more stochasticity in the training process may improve the ability of DAM to perform network pruning. Table \ref{tab:mnist_dam_lenet} (c) further shows the results of DAM on LeNet-5, where we replace all the activation functions with the sine activation function as proposed in \cite{sitzmann2019siren}. Interestingly, the results are much better with sine activation function, implying that the DAM may be compatible with periodic activation functions.

\begin{table}[ht]
\small
\centering
\caption{DAM on MNIST Dataset, LeNet-5 (\# Neurons 6-16-120)}
\label{tab:mnist_dam_lenet}
\begin{subtable}{.99\linewidth}\centering
\subcaption{Tanh Activation w.o. Gradient Noise}
\begin{tabular}{cllllllllll}
\thickhline
$\lambda$ & 0 & 0.01 & 0.02 & 0.05 & 0.1 & 0.2 & 0.3 & 0.4 & 0.5 & 1.0 \\\hline
Parameters & 60.0k & 33.6k & 18.7k & 7.0k & 3.9k & 1.9k & 1.4k & 1.3k & 1.0k & 0.7k \\
Pruned (\%) & 0.00 & 44.15 & 68.82 & 88.41 & 93.59 & 96.78 & 97.71 & 97.85 & 98.31 & 98.90 \\
Accuracy (\%) & 98.85 & 98.61 & 98.47 & 98.16 & 97.49 & 96.54 & 95.47 & 94.79 & 93.70 & 90.12 \\\thickhline
\end{tabular}
\end{subtable}
\begin{subtable}{.99\linewidth}\centering
\subcaption{Tanh Activation with 5\% Gradient Noise}
\begin{tabular}{cllllllllll}\thickhline
$\lambda$       & 0 & 0.01 & 0.02 & 0.05 & 0.1 & 0.2 & 0.3 & 0.4 & 0.5 & 1.0 \\\hline
Parameters      & 60.0k & 35.0k & 18.8k & 5.9k & 3.9k & 1.8k & 1.4k & 1.3k & 1.0k & 0.6k \\
Pruned (\%)     & 0.00 & 41.71 & 68.70 & 90.19 & 93.53 & 96.97 & 97.67 & 97.85 & 98.31 & 99.00 \\
Accuracy (\%)   & 98.78 & 98.57 & 98.69 & 98.03 & 97.69 & 96.61 & 95.64 & 94.78 & 94.16 & 90.16 \\\thickhline
\end{tabular}
\end{subtable}
\begin{subtable}{.99\linewidth}\centering
\subcaption{Sine Activation w.o. Gradient Noise}
\begin{tabular}{cllllllllll}\thickhline
$\lambda$       & 0     & 0.01  & 0.02  & 0.05  & 0.1   & 0.2   & 0.3   & 0.4   & 0.5  & 0.6    \\\hline
Parameters      & 60.0k & 21.5k & 9.2k  & 3.4k  & 1.9k  & 1.5k  & 1.1k  & 1.2k  & 0.8k  & 0.6k  \\
Pruned (\%)     & 0.00  & 64.24 & 84.64 & 94.27 & 96.84 & 97.44 & 98.12 & 98.05 & 98.59 & 98.95 \\
Accuracy (\%)   & 98.93 & 99.00 & 98.85 & 98.53 & 98.38 & 97.12 & 96.12 & 96.63 & 94.91 & 95.49 \\\thickhline
\end{tabular}
\end{subtable}
\end{table}

\subsection{Additional Results on CIFAR Datasets}

Table \ref{tab:cifar} presents additional results of DAM on CIFAR datsets for many other neural network architectures than what was shown in the main paper, including VGG-19, PreResnet-20, PreResNet-56, and PreResnet-110. Note that the DAM results presented in the main paper were obtained using an implementation of our algorithm based on the ChipNet source code \cite{tiwari2021chipnet}, which was published just a few weeks before the time of our submission. Given the recentness of this implementation, we were not able to complete extensive evaluations on all architectures using this implementation. Instead, the results presented in this section are based on an alternate implementation of our DAM algorithm based on the stable source code provided in an older well-established previous work \cite{liu2018rethinking}. In this implementation, the DAM is trained for 160 epochs using AdaBelief \cite{zhuang2020adabelief} optimizer. We release both implementations of our DAM algorithms in the anonymous link of the source code provided in the main paper. 

\begin{table}[ht]
\small
\caption{Additional results of pruning different architectures using DAM on CIFAR datasets.}
\label{tab:cifar}
\centering
\vspace{2ex}
\begin{tabular}{lllllll}\thickhline
Network       & Dataset   & $\lambda$ & Top-1 (\%)  & Params (k) & Pruned C. (\%) & Pruned P. (\%)    \\\hline
VGG-19        & CIFAR-10  & 0.1       & 93.39       & 1,627      &    71.28           &    91.88               \\
VGG-19        & CIFAR-10  & 0.5       & 90.92       & 264        &    89.17           &    98.68               \\
VGG-19        & CIFAR-100 & 0.1       & 73.40       & 6,190      &    41.30           &    69.17               \\
VGG-19        & CIFAR-100 & 0.5       & 67.17       & 664        &    79.81           &    96.69               \\\hline
PreResNet-20  & CIFAR-100 & 0.1       & 67.33       & 228        &    2.72            &    5.88               \\
PreResNet-20  & CIFAR-100 & 0.2       & 66.53       & 217        &    5.37            &    10.50               \\
PreResNet-20  & CIFAR-100 & 0.3       & 66.46       & 206        &    7.72            &    15.13               \\
PreResNet-20  & CIFAR-100 & 0.4       & 65.83       & 191        &    11.25           &    21.18               \\
PreResNet-20  & CIFAR-100 & 0.5       & 66.17       & 183        &    14.12           &    24.67               \\
PreResNet-20  & CIFAR-100 & 1.0       & 63.15       & 121        &    31.84           &    50.05               \\
PreResNet-20  & CIFAR-100 & 2.0       & 54.79       & 47         &    62.35           &    80.45               \\\hline
PreResNet-56  & CIFAR-100 & 0.1       & 72.74       & 573        &    3.33            &    6.60               \\
PreResNet-56  & CIFAR-100 & 0.2       & 72.64       & 526        &    7.93            &    14.35               \\
PreResNet-56  & CIFAR-100 & 0.3       & 72.49       & 493        &    10.72           &    19.67               \\
PreResNet-56  & CIFAR-100 & 0.4       & 72.28       & 466        &    14.62           &    24.14               \\
PreResNet-56  & CIFAR-100 & 0.5       & 71.73       & 419        &    19.17           &    31.73               \\
PreResNet-56  & CIFAR-100 & 1.0       & 69.94       & 314        &    33.60           &    48.77               \\
PreResNet-56  & CIFAR-100 & 2.0       & 64.31       & 153        &    62.94           &    75.05               \\\hline
PreResNet-110  & CIFAR-10 & 0.1       & 94.51       & 975    	 &    9.72       	  &    15.01               \\
PreResNet-110  & CIFAR-10 & 0.2       & 93.73       & 808    	 &    20.71       	  &    29.54               \\
PreResNet-110  & CIFAR-10 & 0.3       & 93.28       & 646    	 &    31.10       	  &    43.65               \\
PreResNet-110  & CIFAR-10 & 0.4       & 93.35       & 560    	 &    37.08       	  &    51.18               \\
PreResNet-110  & CIFAR-10 & 0.5       & 92.97       & 482    	 &    46.55       	  &    58.00               \\
PreResNet-110  & CIFAR-10 & 1.0       & 91.61       & 208    	 &    72.51       	  &    81.85               \\
PreResNet-110  & CIFAR-10 & 2.0       & 88.90       & 89    	 &    84.99       	  &    92.21               \\\hline
PreResNet-110  & CIFAR-100 & 0.1      & 75.55       & 1110    	 &    4.20       	  &    5.15               \\
PreResNet-110  & CIFAR-100 & 0.2      & 74.73       & 1010    	 &    10.24       	  &    13.70               \\
PreResNet-110  & CIFAR-100 & 0.3      & 75.08       & 940    	 &    14.54       	  &    19.63               \\
PreResNet-110  & CIFAR-100 & 0.4      & 73.93       & 873     	 &    18.37       	  &    25.41               \\
PreResNet-110  & CIFAR-100 & 0.5      & 73.44       & 817    	 &    21.87       	  &    30.17               \\
PreResNet-110  & CIFAR-100 & 1.0      & 71.63       & 547    	 &    39.68       	  &    53.21               \\
PreResNet-110  & CIFAR-100 & 2.0      & 68.32       & 261    	 &    66.45       	  &    77.72               \\
\thickhline
\end{tabular}
\end{table}

\subsection{Analyzing Similarity in Pruned Features}
\begin{figure}[H]
\begin{subfigure}{0.24\textwidth}
    \centering
    \includegraphics[width=0.90\linewidth]{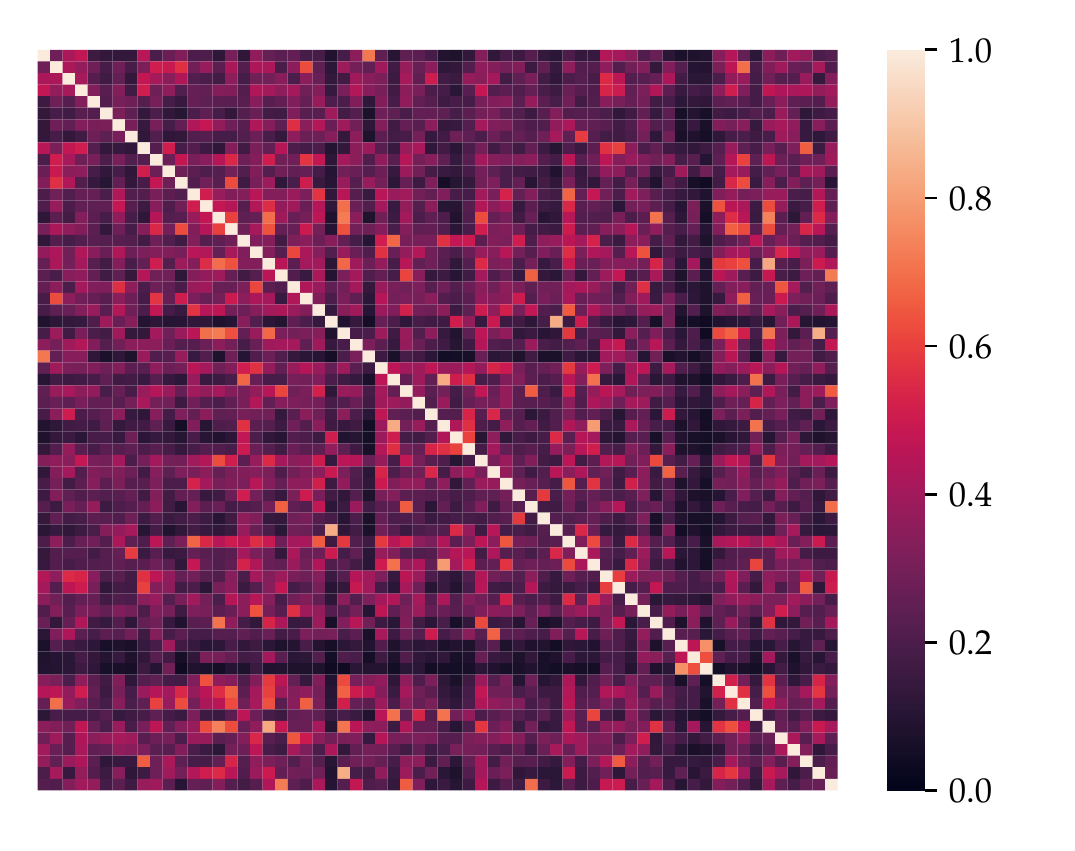}
    \caption{Unpruned}
    \label{fig:cka_unpruned}
\end{subfigure}
\begin{subfigure}{0.24\textwidth}
    \centering
    \includegraphics[width=0.90\linewidth]{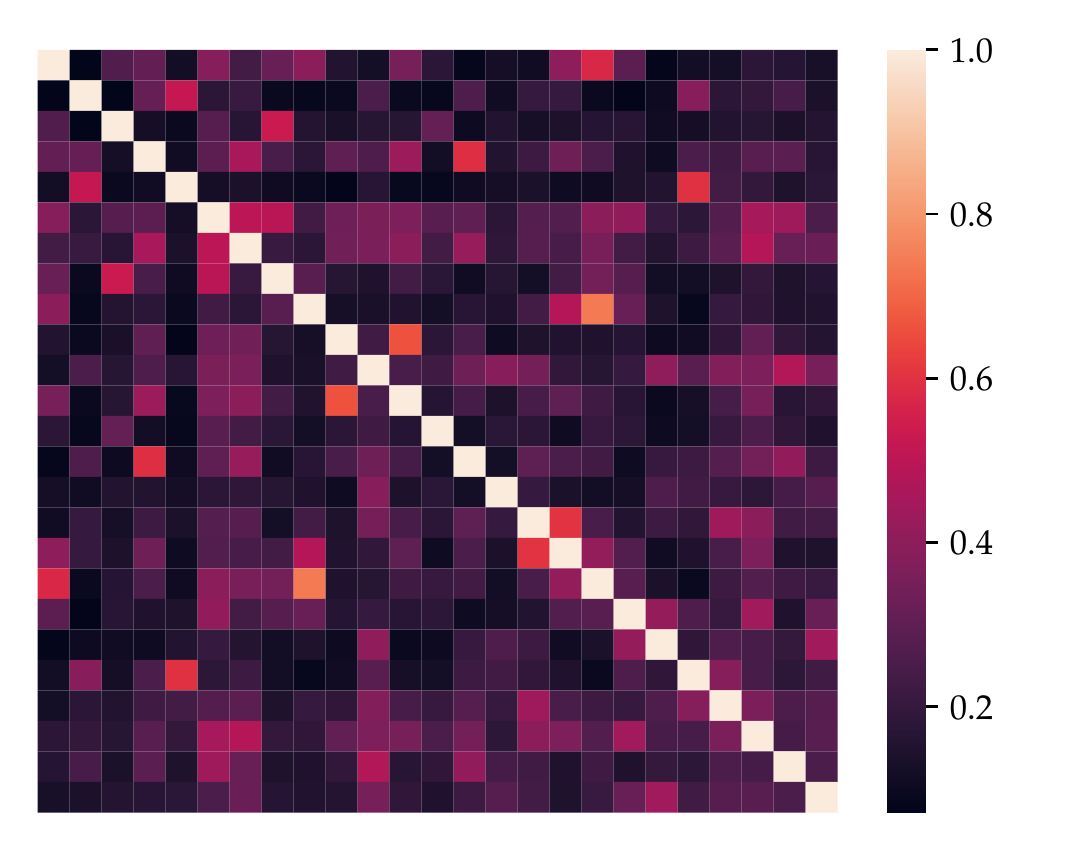}
    \caption{DAM}
    \label{fig:cka_dam}
\end{subfigure}
\begin{subfigure}{0.24\textwidth}
    \centering
    \includegraphics[width=0.99\linewidth]{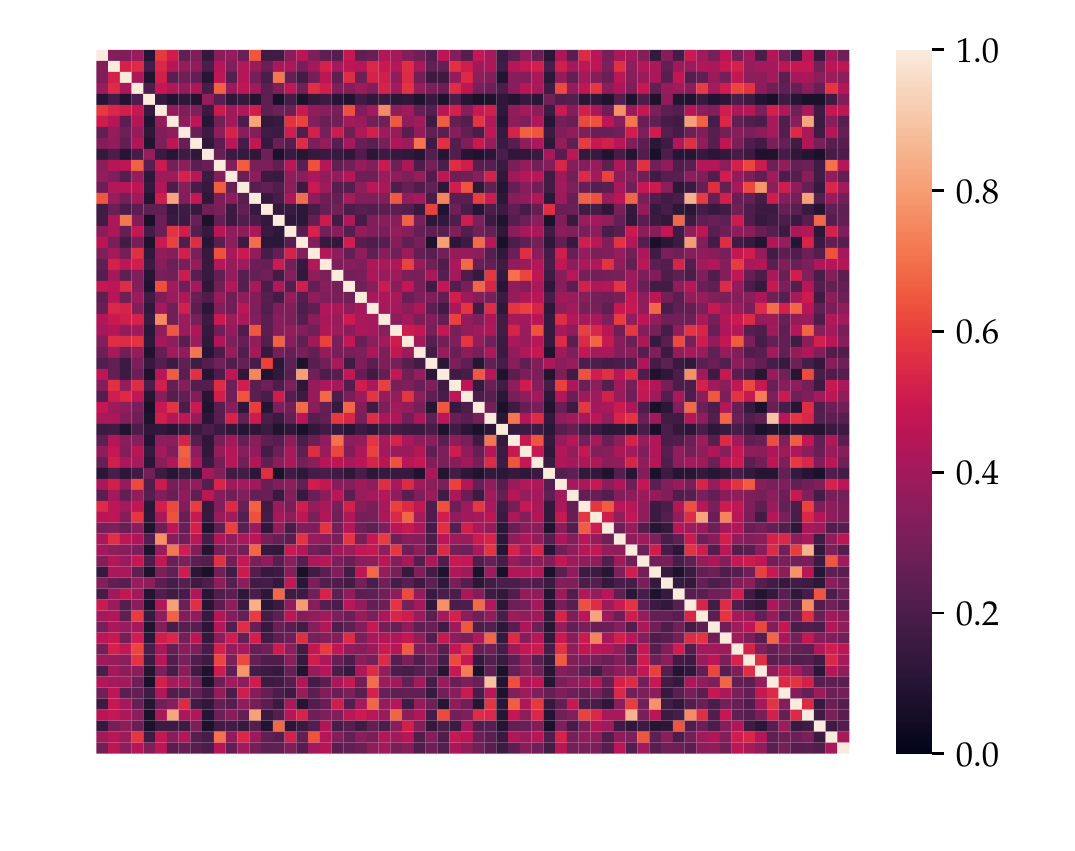}
    \caption{Net-Slim}
    \label{fig:cka_netslim}
\end{subfigure}
\begin{subfigure}{0.24\textwidth}
    \centering
    \includegraphics[width=0.99\linewidth]{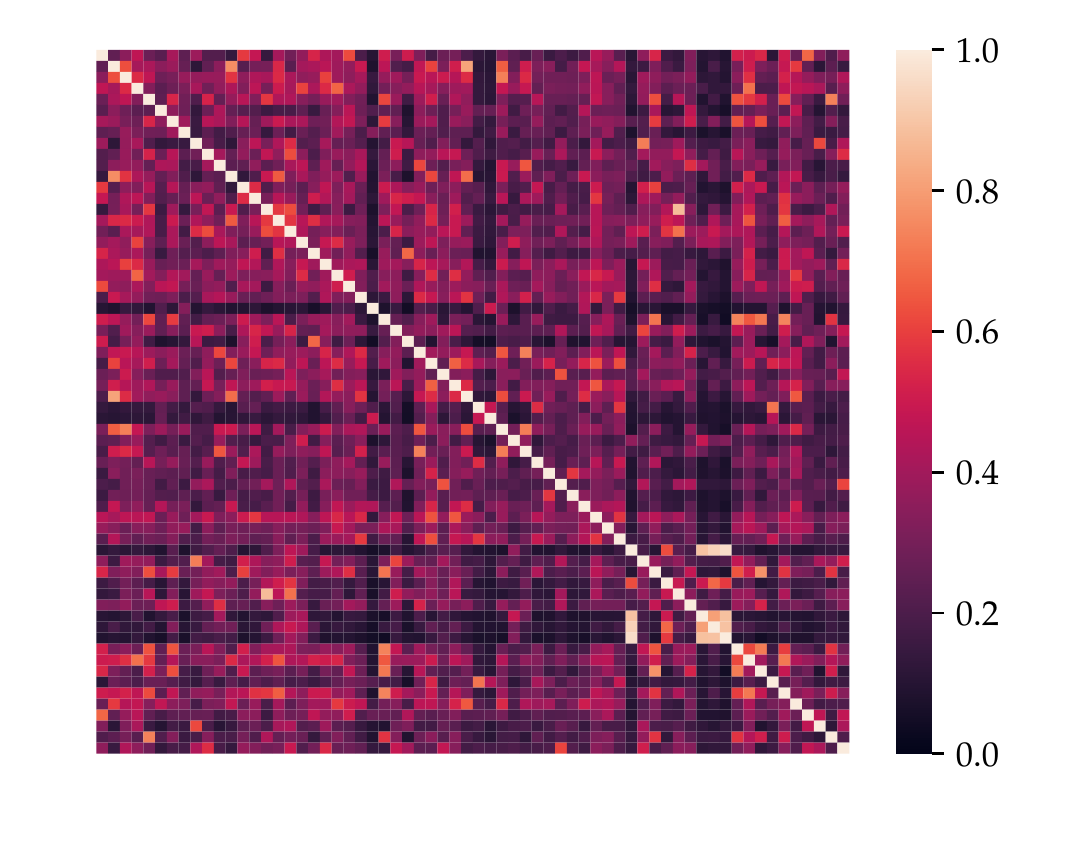}
    \caption{ChipNet}
    \label{fig:cka_chipnet}
\end{subfigure}
\caption{CKA similarity between features learned at layer 54 of PreResNet-164 model, before pruning, and after pruning using the three methods.}
\label{fig:cka}
\end{figure}

A useful property of \textit{compact} features learned at the hidden layers of a neural network is that they show express `distinct' features upon convergence, such that pruning the neurons any further would lead to drop in accuracy. To evaluate the similarity in the features extracted by comparative structured network pruning methods, we compute the centered kernel alignment (CKA) similarity \cite{kornblith2019similarity} matrix for all pairs of neurons with non-zero mask values at the 54-th layer in the PreResNet-164 architecture pruned using DAM, Net-Slim, and ChipNet. Note that we chose the 54-th layer as it represents one third of the total number of layers at the end of the first `BaseBlock' of PreResNet-164 for ease of implementation, although these results can be visualized for any other layer number too. The CKA similarity matrices of the unpruned network and the three pruned networks are shown in Figure \ref{fig:cka}, where higher off-diagonal values in these matrices represent higher similarity among the features. To further quantify the differences between DAM and the baseline methods, we compute the statistics of the values in the off-diagonal elements of the CKA matrices in Table \ref{tab:cka}. We can observe observe that DAM shows lowest CKA similarity on the off-diagonal elements as compared to Net-Slim and ChipNet, indicating that the features learned by DAM are more distinct from one another. Also note that in Figure \ref{fig:cka}, the size of the pruned features (with non-zero values) extracted by DAM is quite smaller than what we obtain from Net-Slim and ChipNet using their standard implementations made available by their authors. This is because some proportion of pruned channels become nonzero after finetuning in Net-Slim (see details of mask implementation here\footnote{ \url{https://github.com/Eric-mingjie/network-slimming/tree/master/mask-impl}}) and ChipNet. Their current implementations thus result in smaller actual pruning ratios after finetuning for practical use than what is reported (before finetuninng).

\begin{table}[ht]
\small
\caption{Statistics of CKA similarity between different features learned at layer 54 of PreResNet-164 after pruning (calculated using the off-diagonal elements in Figure \ref{fig:cka}).}
\label{tab:cka}
\centering
\vspace{2ex}
\begin{tabular}{llllll}\thickhline
CKA   & Unpruned  & DAM     & Net-Slim  & ChipNet  \\\hline
mean  & 0.273     & \textbf{0.229}   & 0.328     & 0.295    \\
std   & 0.018     & \textbf{0.013}   & 0.021     & 0.020    \\
max   & 0.840     & \textbf{0.740}   & 0.893     & 0.964    \\
\thickhline
\end{tabular}
\end{table}

\subsection{Additional Results on TinyImageNet Datasets}

\begin{table}[H]
\small
\caption{Additional results of pruning PreResNet-164 using DAM on TinyImageNet datasets.}
\label{tab:tinyimagenet}
\centering
\vspace{2ex}
\begin{tabular}{llllllll}\thickhline
$\lambda$      & 0.0   & 0.1   & 0.3   & 0.5   & 0.7   & 1.0   & 2.0   \\ \hline
Top-1 (\%)     & 52.72 & 52.75 & 52.51 & 52.99 & 52.11 & 52.89 & 53.17 \\
Pruned C. (\%) & 0.00  & 6.03  & 20.05 & 30.91 & 38.71 & 45.94 & 68.52 \\
\thickhline
\end{tabular}
\end{table}

We managed to verify the performance of DAM on a larger dataset - TinyImageNet. For this dataset, we used PreResNet-164 as the backbone network with standard configurations of DAM as CIFAR-100 (see Appendix D.2). We used the AdaBelief \cite{zhuang2020adabelief} optimizer for faster convergence on this larger dataset. The optimizer configurations are the same as used for
the results shown in Table \ref{tab:tinyimagenet}. We can see that DAM is able to consistently maintain the test accuracy on TinyImageNet even at extreme levels of network pruning close to 68\%. This demonstrates the ability of DAM to achieve useful generalization on larger datasets going beyond CIFAR. 

\end{document}